\documentclass[fontsize=9pt]{scrartcl}
\usepackage{graphicx}
\usepackage{amssymb}
\usepackage{amsmath}
\usepackage{algorithm}
\usepackage[noend]{algpseudocode}
\usepackage{url}
\usepackage{diagbox}
\usepackage[super, sort&compress]{natbib}
\usepackage[colorlinks=true,citecolor=black,linkcolor=black,urlcolor=black]{hyperref}
\usepackage[a4paper, top=1.9cm, bottom=1.5cm, left=2cm, right=2cm, includefoot]{geometry}
\usepackage[affil-it]{authblk}
\usepackage[multiple]{footmisc}
\usepackage{marvosym}

\newcommand{\tf}{\texttt{tensorflow}}

\newcommand{\R}{\mathbb{R}}

\let\svthefootnote\thefootnote
\newcommand\freefootnote[1]{%
  \let\thefootnote\relax%
  \footnotetext{#1}%
  \let\thefootnote\svthefootnote%
}

\DeclareUrlCommand\UScore{\urlstyle{rm}}
\newcommand{\doi}[1]{\href{https://doi.org/#1}{\UScore{#1}}}
\renewcommand{\url}[1]{\href{#1}{\UScore{#1}}}

\title{ECToNAS: Evolutionary Cross-Topology Neural Architecture Search}
\author[1]{Elisabeth J. Schiessler\textsuperscript{$*$,\Letter,}}
\author[1,2]{Roland C. Aydin\textsuperscript{$*$,\Letter,}}
\author[1,2]{Christian J. Cyron}
\affil[1]{Institute of Material Systems Modeling, Helmholtz-Zentrum Hereon, Max-Planck-Strasse 1, 21502 Geesthacht, Germany}
\affil[2]{Institute for Continuum and Material Mechanics, Hamburg University of Technology, Eissendorfer Strasse 42, 21073 Hamburg, Germany}

\date{February 2024}

\begin{document}

\maketitle
\freefootnote{$^*$ \textit{equal contribution}}
\freefootnote{\Letter~ elisabeth.schiessler@hereon.de, roland.aydin@hereon.de}
\begin{abstract}
We present ECToNAS, a cost-efficient evolutionary cross-topology neural architecture search algorithm that does not require any pre-trained meta controllers.
Our framework is able to select suitable network architectures for different tasks and hyperparameter settings, independently performing cross-topology optimisation where required.
It is a hybrid approach that fuses training and topology optimisation together into one lightweight, resource-friendly process.
We demonstrate the validity and power of this approach with six standard data sets (CIFAR-10, CIFAR-100, EuroSAT, Fashion MNIST, MNIST, SVHN), showcasing the algorithm's ability to not only optimise the topology within an architectural type, but also to dynamically add and remove convolutional cells when and where required, thus crossing boundaries between different network types.
This enables researchers without a background in machine learning to make use of appropriate model types and topologies and to apply machine learning methods in their domains, with a computationally cheap, easy-to-use cross-topology neural architecture search framework that fully encapsulates the topology optimisation within the training process.
\end{abstract}

\section{Introduction}
\label{sec:Introduction}
Deep learning as a tool to understand complex relationships within data sets and make predictions on unseen data has been around since the advent of machine learning.
In recent years it has become more and more popular not only in a huge number of scientific fields but also for example in commercial applications, which in turn again attracts more people to the idea that making use of deep learning can be beneficial for them.
Yet while it may be becoming easier to get started on deep learning, actually using it to achieve `good' results can hinge on a variety of factors.

One of these crucial aspects is selecting an adequate network architecture, which also means deciding which types of architectural elements should be included.
The optimal choice may be highly dependent on the specific task and data set, and even with expert deep learning knowledge finding said optimum can require hours and hours of computation time.
Especially in applied sciences there are often limits to the available computational budget for deep learning.
Additional difficulties may include an unclear task definition, a limited amount of training data, or data that is of mixed types (e.g. image data combined with additional measurements in medical applications).

Neural architecture search is a specialised branch of automatic machine learning (auto ML) that aims to mitigate some of these problems by limiting the required amount of human expert knowledge.
It can be applied to independently find the ideal network architecture given the specified task and available data.

This work extends our previously published method called \emph{the Surgeon} \citep{Schiessler2021}, which introduced a resource-friendly neural architecture search algorithm for multilayer perceptrons, also known as feed forward neural networks (FFNNs), that works especially well on restricted computing resources and limited training data.
We combine it with network growth operations from \citet{Chen2016} and state of the art, resource-friendly structured pruning techniques based on \citet{Liu2020}.
The result is ECToNAS, a cost-efficient evolutionary cross-topology neural architecture search algorithm that works as a stand alone method and in particular does not require pre-training any high level selector algorithms or meta controllers.
It re-uses network weights between different candidates, and integrates training of the final architecture within the search process, thus reducing the hypothetically required amount of training time by around 80\% when compared to re-training each candidate network from scratch.

We provide two different modes that can be applied as per user requirements.
Standard ECToNAS focuses on maximising the specified target metric (such as validation accuracy), and manages to outperform the naive baseline throughout all our experiments.
By setting a so-called greediness parameter it is possible to influence ECToNAS's scoring function such that more compressed network architectures might get favoured even at the cost of a small reduction in target metric.
With this mode we are able to shrink network parameter counts by up to 90\% while losing only around 5-7 percentage points (or less) in target metric results compared to the greedy version.

\begin{figure}[htb]
    \centering
    \includegraphics[width=\textwidth]{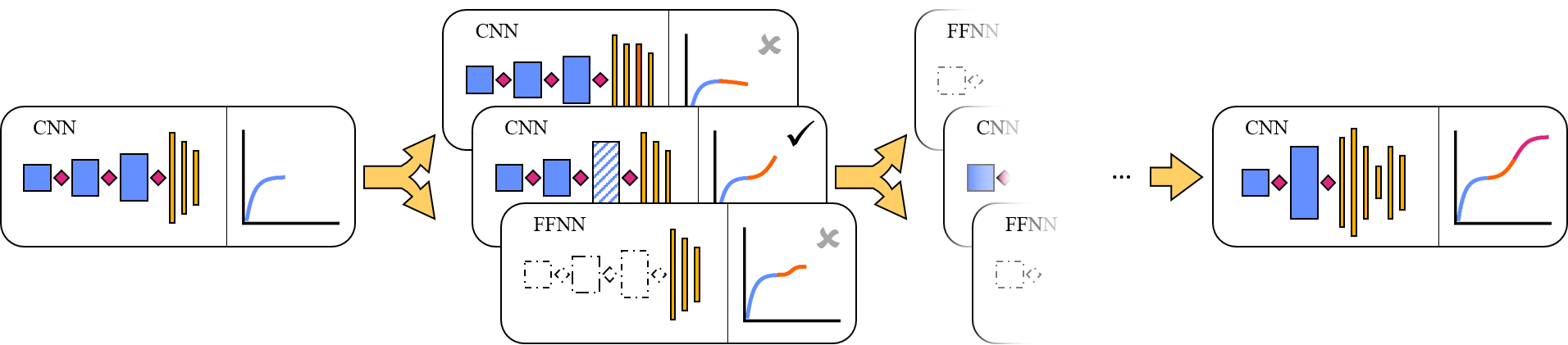}
    \caption{Graphical representation of the ECToNAS algorithm}
    \label{fig:graphical_abstract}
\end{figure}

ECToNAS is able to select suitable network architectures for different tasks and hyperparameter settings, and independently performs cross topology optimisation where required.
By this we mean that starting from one topology type, such as a convolutional neural network (CNN), the combined training and topology optimisation may result in a different type of network, such as a standard FFNN, and vice versa.

The underlying code can be accessed via \url{https://github.com/ElisabethJS/ECToNAS}.

In this work we showcase the possibilities of re-using network weights in conjunction with cross topology optimisation and an evolutionary algorithm, and present some primers for further studies.

\section{Related Works}\label{sec:related}
Early versions of neural architecture search (NAS) have been around since at least \citet{Tenorio1988}.
The idea of automated NAS for deep learning has been revitalised with groundbreaking works by \citet{Zoph2017} and \citet{Liu2019}, which have seen a huge number of extensions in various directions \citep{Ren2021}, often relying on weight sharing between different network candidates \citep{Pham2018, Dong2019, Bender2020, Yu2020, Xu2020, Xie2021, Li2021}.

Evolutionary algorithms that continuously optimize and evolve network topologies seem to be a promising approach \citep{Zhou2021}.
However while reproduction-style algorithms such as \citet{Xue2021} might be quite successful in their use of ResNet \citep{He2016} and DenseNet \citep{Huang2017} blocks, they rely on re-training from scratch an exponential amount per generation (23 GPU Days for CIFAR-10).
More resource-aware approaches rely on incorporating pruning into the training process \citep{Peng2022} or departing from the traditional two-stage setup, where first the optimal architecture is identified and then it is retrained from scratch and evaluated \citep{Yu2020}.

Pruning techniques can be roughly categorised into unstructured or structured approaches \citep{Blalock2020}.
Unstructured pruning regards each connection within a neural network individually and identifies those that may be set to zero, leaving the overall sequence of layers and their topological specifics intact.
The majority of publications on pruning techniques focuses on unstructured pruning which can show great results by using methods such as learning rate rewinding \citep{Le2021} or knowledge extraction \citep{Mirkes2020}.
Many attempts are made at reducing the required amount of retraining \citep{Liu2020, Gurevin2021, Tian2021},
with some focussing on various low-cost approximations and single shot estimators \citep{Lee2019, Wang2020, Tanaka2020, Arnob2021, Chen2021, Siems2021, Miao2022}.

Structured pruning on the other hand aims to identify topological elements that can be removed completely.
While there are some structured pruning approaches that are applied post training \citep{Liu2017, Idelbayev2021}, others introduce additional regularization terms that promote desired sparsification already during training \citep{Wen2016, Huang2022}.
As an intermediate between structured and unstructured pruning, \citet{Zhang2022} uses filters to align network elements.

NAS for image classification traditionally almost exclusively focuses on achieved validation accuracy as the target metric for deciding between various available model structures.
In recent years, loss based evaluation metrics have been successfully applied especially when comparing model proxies instead of fully trained models \citep{Lee2019, Wang2020, Tanaka2020, Abdelfattah2021}, with \citet{Ru2021} concluding that loss might in general be favourable over accuracy for candidate evaluation.
None of the above approaches directly consider the number of network parameters in their decision metric, which can lead to quite large and over parameterised final architectures \citep{Larsen2022}.
Some works penalise network size through different strategies, such as including some form of efficiency term within their loss function \citep{Peng2022}.

A truly merged cross-architecture NAS system such as ECToNAS has hitherto not been proposed to the best of our knowledge.

\section{Methods}\label{sec:methods}
We begin by explaining the underlying methods and techniques that ECToNAS is built upon.
The steps for manipulating multilayer perceptrons which were introduced in \emph{the Surgeon} \citep{Schiessler2021} and are used without further refinement or modification are briefly summarised in Appendix \ref{app:fully_connected}.
A graphical representation of the ECToNAS algorithm is provided in Figure \ref{fig:graphical_abstract}.

\subsection{Genetic Neural Architecture Search}
Genetic algorithms (also called evolutionary algorithms) are inspired by natural processes such as evolution and mutation.
A population of individuals competes towards some predefined goal, and the fittest members are able to reproduce and pass on (some of) their defining elemental structures.
ECToNAS is a genetic algorithm that cyclically generates child populations $C_i$ of neural network architecture candidates based off a number of parent networks $P_i$ using mutation operations.
A high level code summary is provided in Algorithm \ref{alg:ectonas}.

\begin{algorithm}[htb]
    \caption{ECToNAS}
    \label{alg:ectonas}
    \begin{algorithmic}[1]
        \Function{Run}{starting topology $t_0$}
            \State pre-train $t_0$ (warm start)
            \State initialise parent generation $P_0 = \{t_0\}$
            \While{termination criteria not met}
                \State \# Mutation phase
                \For{\textbf{each} individual $p$ \textbf{in} parent generation $P_i$}{}%\Comment{mutation phase}
                    \State create potential offspring $o_p$, reject degenerates
                \EndFor
                \State offspring from all parents form child generation $C_i = \bigcup_{p \in P_i}{o_p}$
                \State
                \State \# Competition phase 1
                \State determine best competitor per operation type%\Comment{competition phase 1}
                \State
                \State \# Competition phase 2
                \State determine best remaining children $C_i^*$ from all types%\Comment{competition phase 2}
                \State
                \State \# Evolution phase
                \State winning children become new parent generation $P_{i+1} = C_i^*$%\Comment{evolution phase}
            \EndWhile
            \State perform final selection to determine optimised topology $t_{final}$
            \State \Return{$t_{final}$}
        \EndFunction
    \end{algorithmic}
\end{algorithm}

These mutation operations not only change the network architecture (by adding or reducing neurons, channels or even whole layers), but also manipulate network weights in such a way as to incur minimal possible change in input-output behaviour between a parent and its modified offspring.
This means adjusting network weights to fit to new shapes while preserving the maximal amount of information possible.
The child generation then runs through two bracket style competitions in order to select the fittest individuals, which in turn become the new parent generation.
Training occurs during the competition phases.
The algorithm terminates when the allowed computational budget is spent.

We split selecting the fittest candidates into two phases in order to increase fairness during the selection process.
More drastic changes will usually come at the cost of a somewhat reduced accuracy score, for which the network candidates need some training time to recuperate.
Furthermore, smaller networks are able to converge faster than larger ones.
Thus at first we only compare children that have undergone the same type of mutation, which all should have roughly the same pace of convergence.

In phase 1 of the competitions, only children of the same mutation type (for example adding channels to a convolutional layer) compete against each other, producing one winner per mutation type.
Each pool of competitors is divided into pairs of 2 candidates. These receive a few epochs of training, then the one with the lower fitness score is eliminated from the pool.

In phase 2, all remaining offspring compete against each other, producing $n$ winners $C_i^*$ that in turn form the new parent generation $P_{i+1}$. 
$n$ is a hyperparameter that can be set by the user.
Higher $n$ potentially yields better results as more individuals are retained in the parent generation, thus allowing ECToNAS to search a larger population of candidates, which in turn also increases the computational cost of the algorithm.

\begin{figure}[htb]
    \centering
    \begin{minipage}{\textwidth}
    Competition phase 1 \hfill
    \end{minipage}
    \begin{minipage}{0.31\textwidth}
    \includegraphics[width=\textwidth]{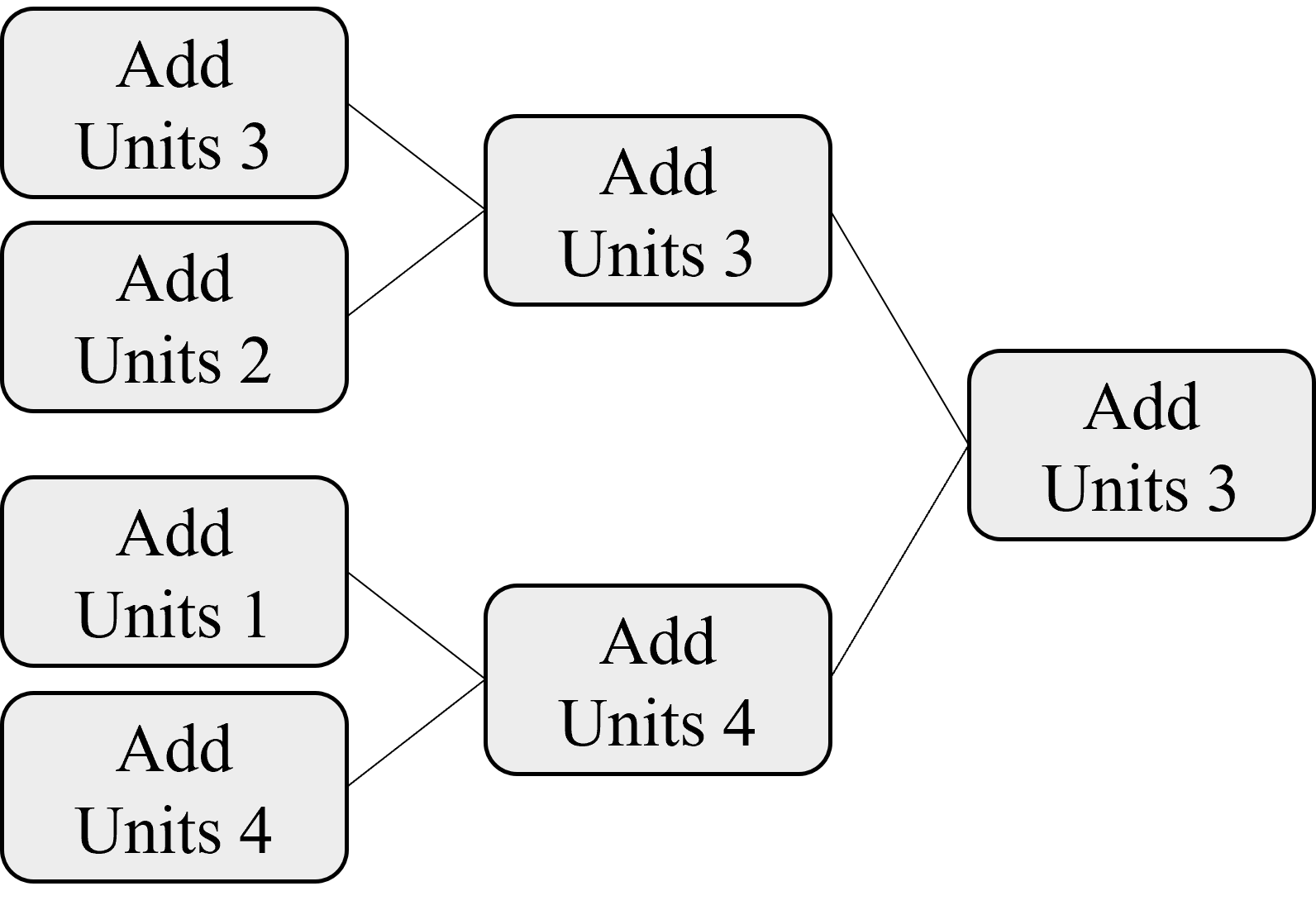}
    \end{minipage}\hfill
    \begin{minipage}{0.31\textwidth}
    \includegraphics[width=\textwidth]{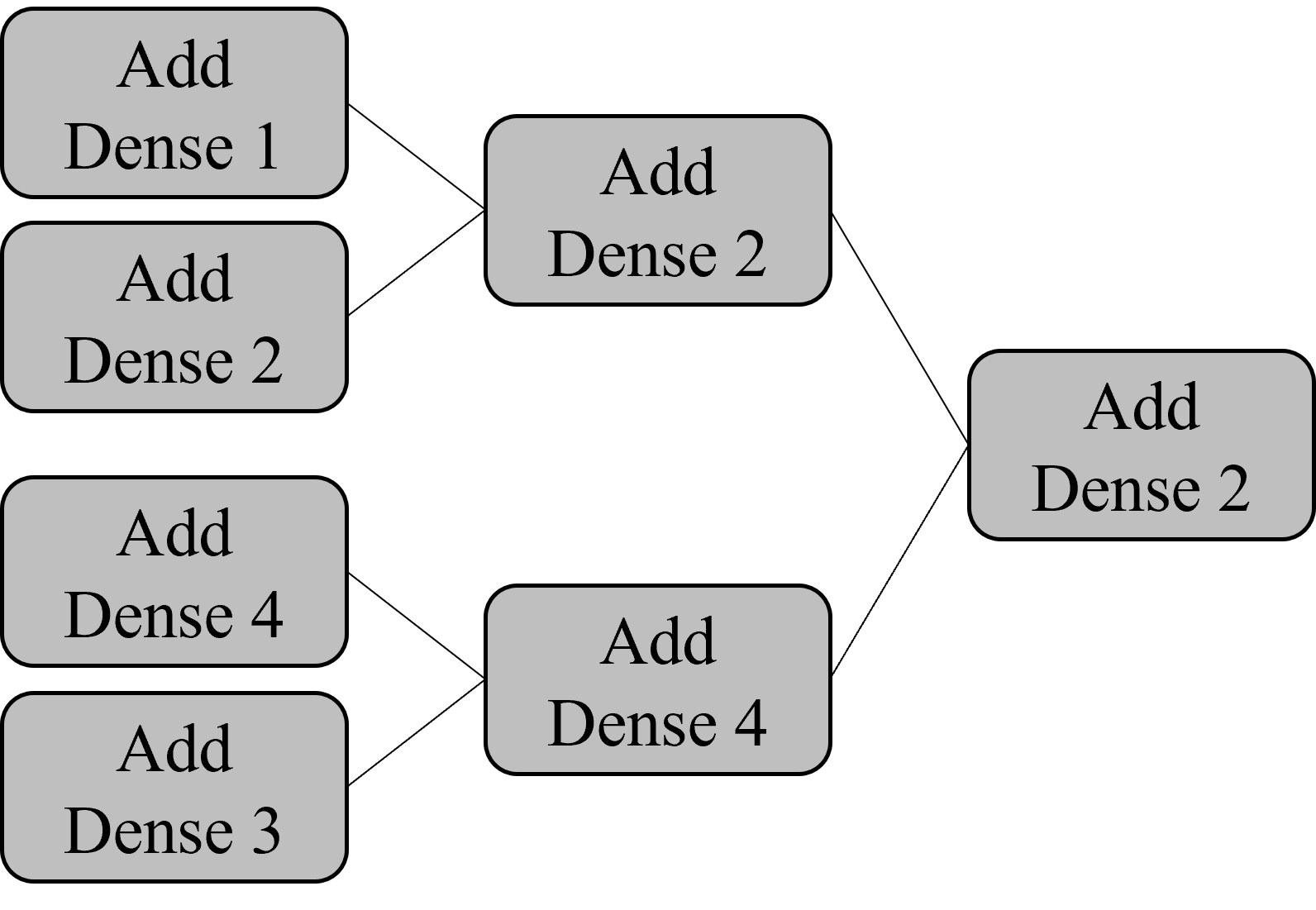}
    \end{minipage}\hfill
    \begin{minipage}{0.31\textwidth}
    \includegraphics[width=\textwidth]{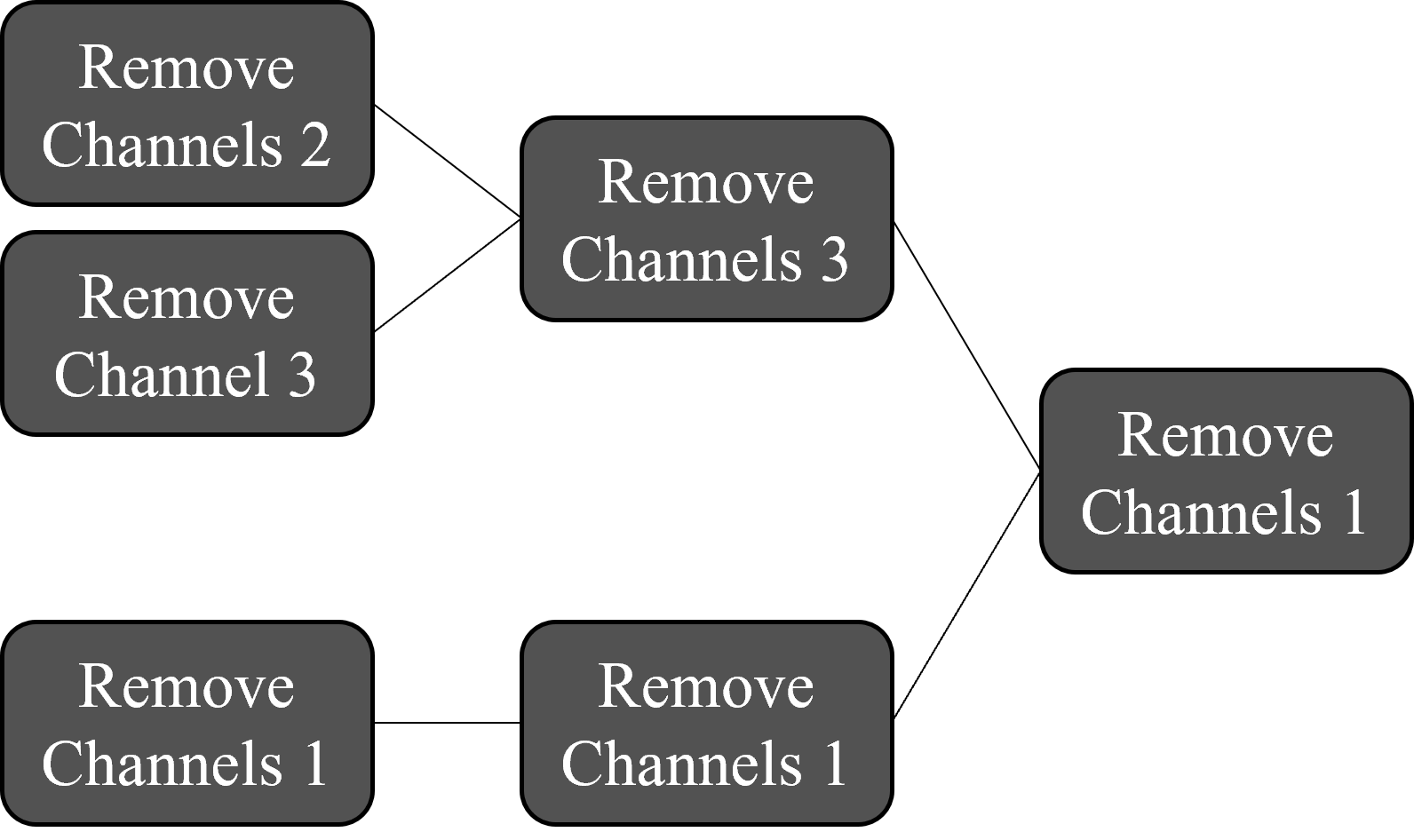}
    \end{minipage}
    \begin{minipage}{0.31\textwidth}
    \begin{minipage}{\textwidth}
    Competition phase 2 \hfill
    \end{minipage}
    \includegraphics[width=\textwidth]{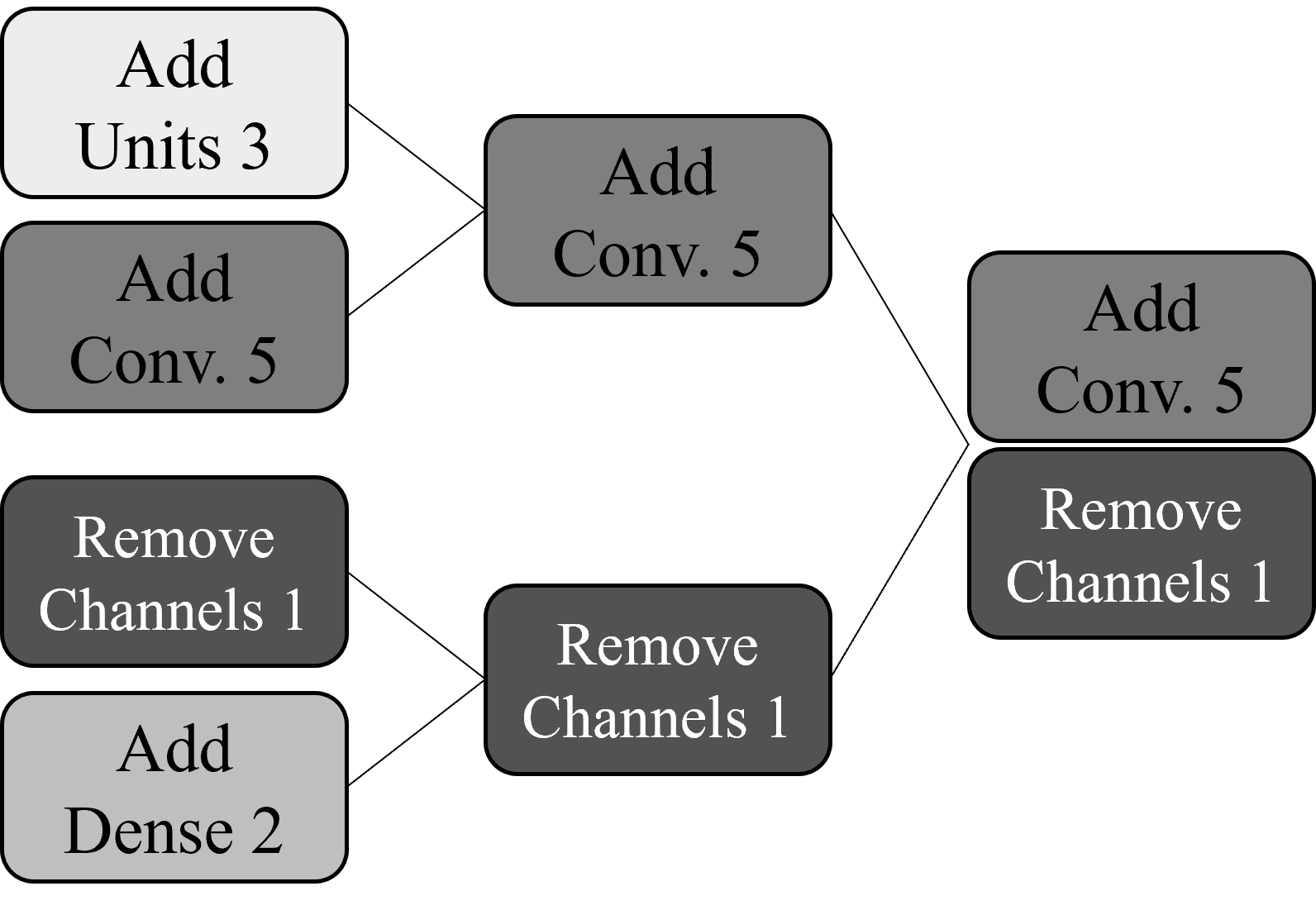}
    \end{minipage}\hfill
    \begin{minipage}{.65\textwidth}
    \caption{Bracket style competitions.\\
    Top: In phase 1, one winning candidate is selected per mutation type. Offspring are randomly paired into groups of two out of their assigned pools, lucky tickets occur in case of an odd number of competitors.\\ Bottom left: In phase 2, $n$ overall winners are selected from phase 1 winners. Here, $n=2$. These will form the next parent generation.}
    \end{minipage}
    \label{fig:competition}
\end{figure}

Training phases are already integrated into the competition process, and generating offspring re-uses already learned network weights, thus greatly reducing the required amount of training.
Compared to \emph{the Surgeon} \citep{Schiessler2021}, ECToNAS requires less overall training time and can access a much larger search space.
Next to fully connected and various adaptor layers, ECToNAS's search space also includes convolutional, pooling, as well as batch normalisation layers, thus also allowing different topologies.
It is able to independently decide which architecture type is best suited for the provided training data and task, and can add or remove structural elements as required and even change the type of network topology altogether.

\subsection{Available Network Operations}
During the mutation phase of Algorithm \ref{alg:ectonas}, ECToNAS has a number of modification operations available to generate mutated offspring from any given parent network.
By design, these modifications change the network architecture as well as relevant weights, such that the overall input-output behaviour changes as little as possible.
\citet{Chen2016} have shown that under piecewise linear activation functions (such as \texttt{relu}), adding units to dense layers, channels to convolutional layers or even whole convolutional or dense layers can be performed with zero change to the overall network output.
In \citet{Schiessler2021}, we demonstrated how to use singular value decomposition (SVD) to remove units from dense layers, or remove whole dense layers with as little information loss as possible.
Summaries of these operations are included in Appendix \ref{app:fully_connected}.

\begin{figure}[htb]
    \centering
    \begin{minipage}{.4\textwidth}
    \includegraphics[width=\textwidth]{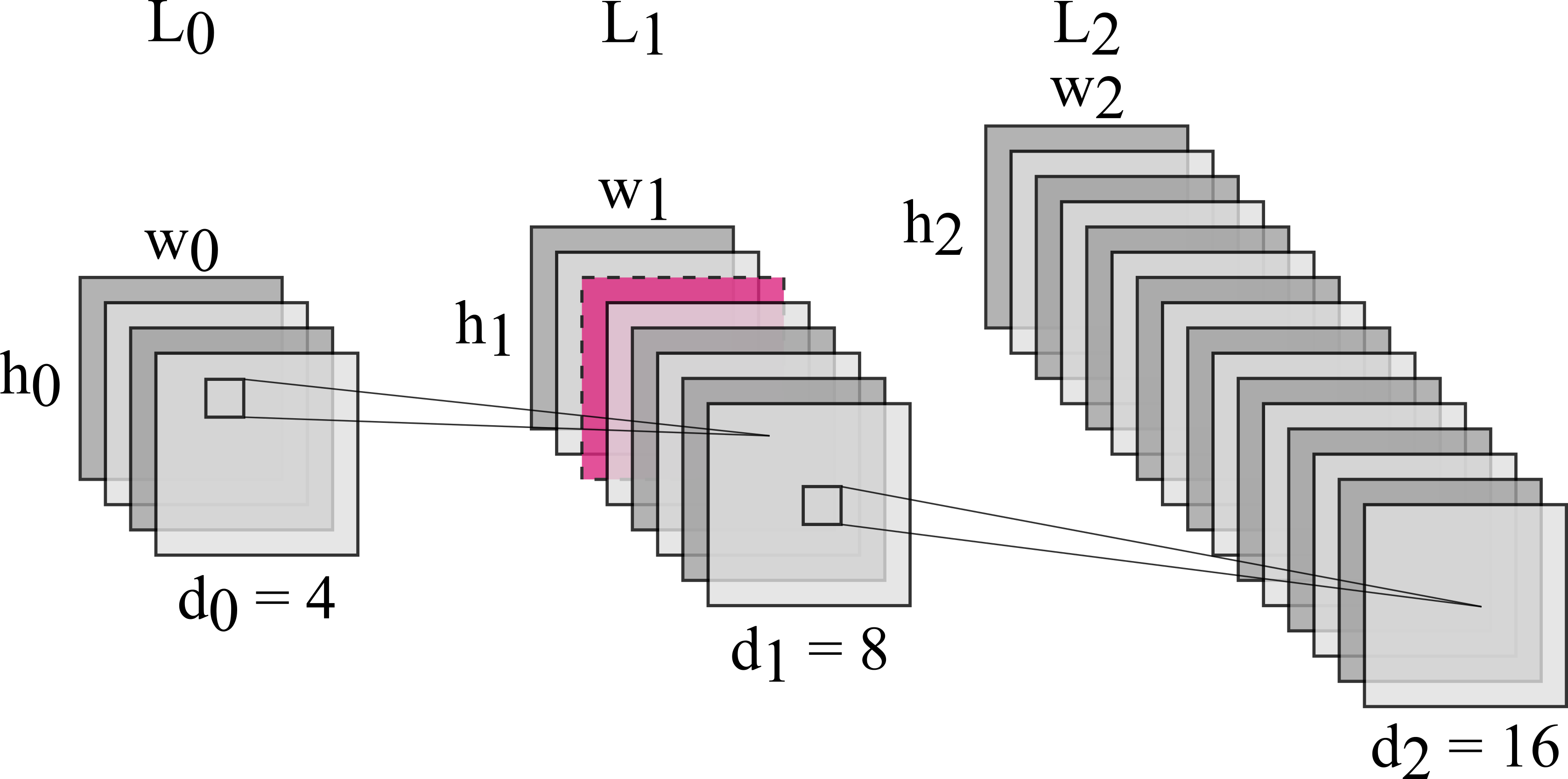}
    \end{minipage}\hfill
    \begin{minipage}{.28\textwidth}
    \includegraphics[width=\textwidth]{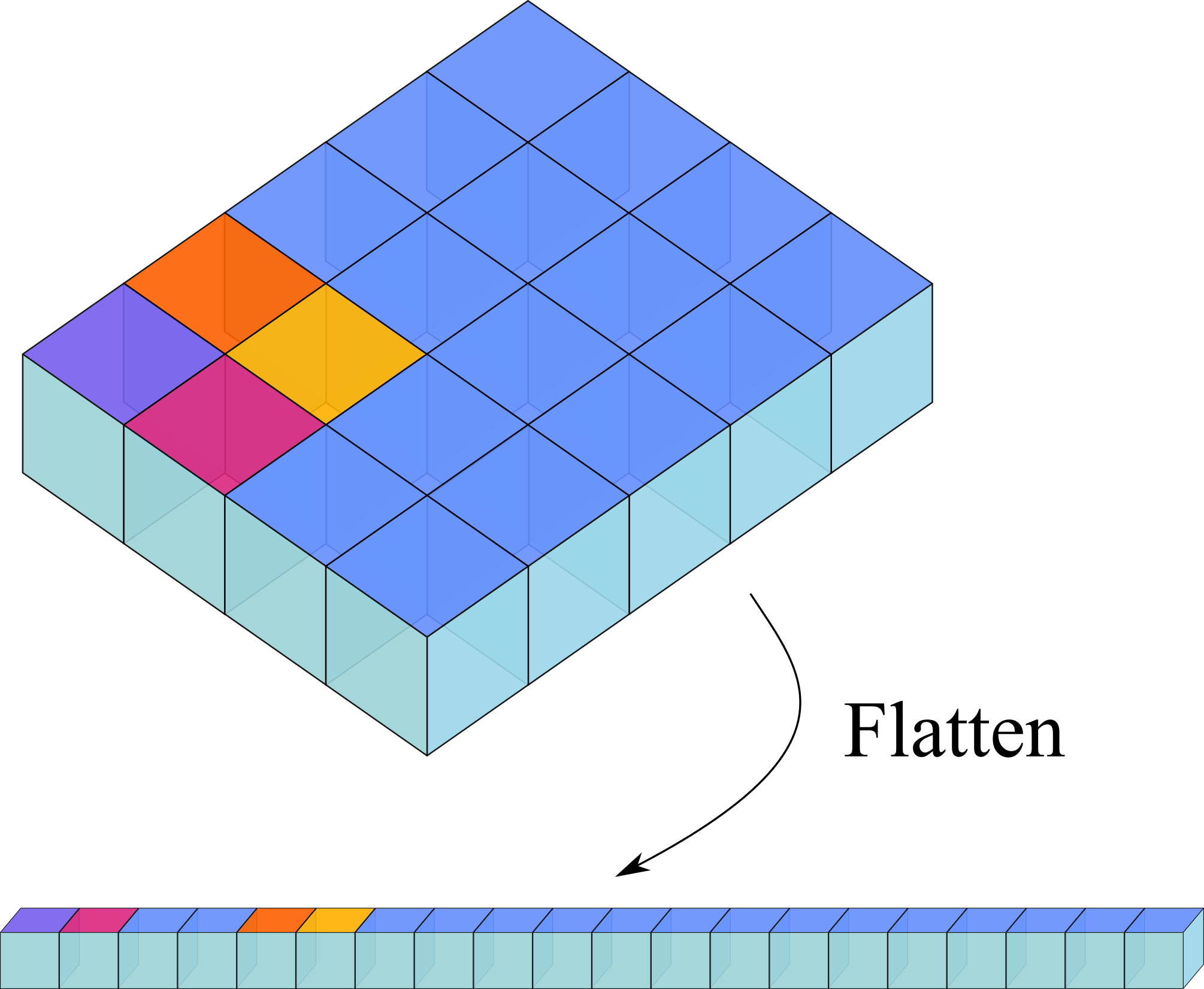}
    \end{minipage}\hfill
    \begin{minipage}{.28\textwidth}
    \includegraphics[width=\textwidth]{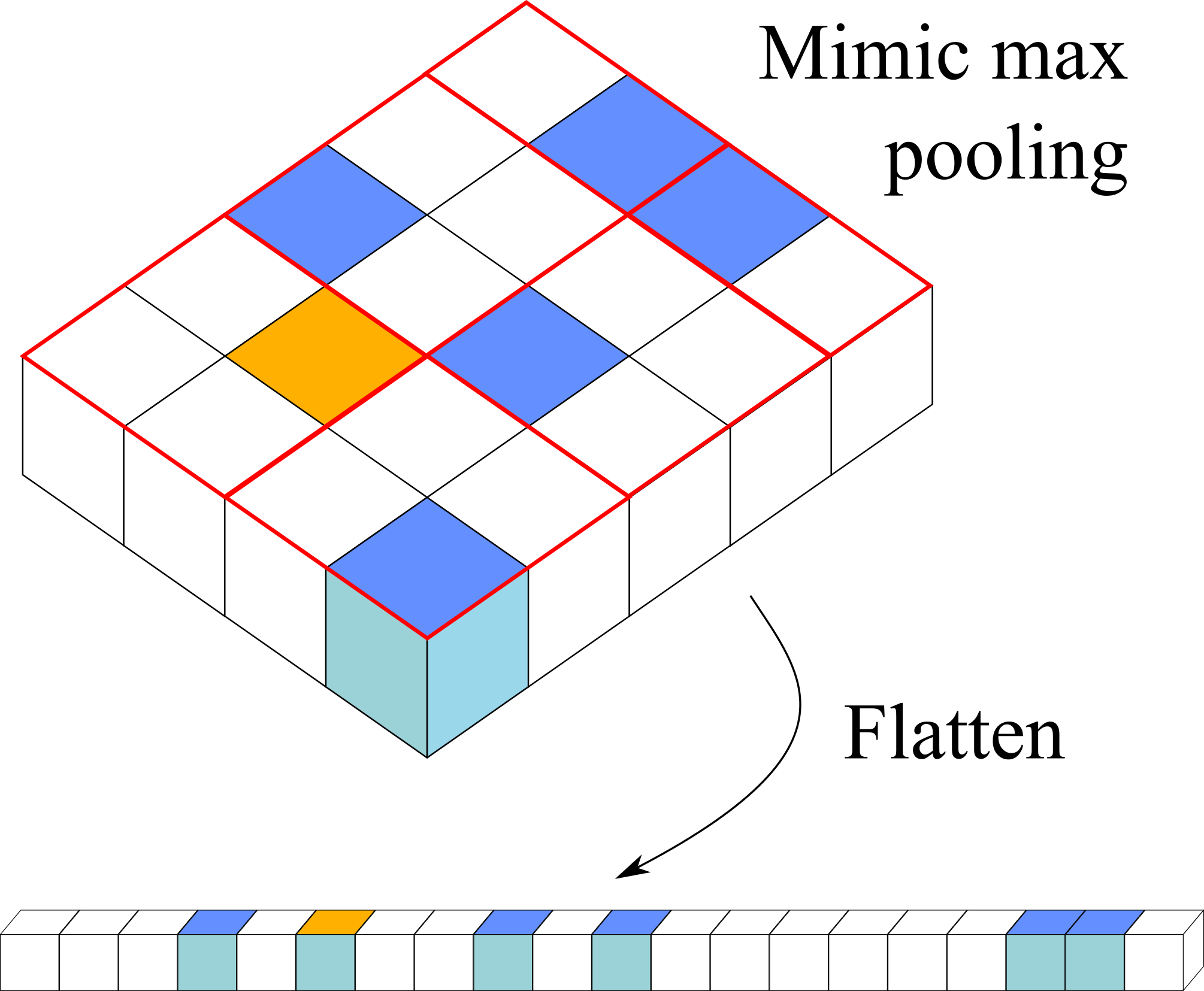}
    \end{minipage}
    \begin{minipage}[t]{.4\textwidth}
     \caption{Example of a three layer CNN.
     Removing a channel from $L_1$ (coloured purple) affects all incoming and outgoing connections of that layer. Created using \citet{LeNail2019}}
    \label{fig:cnn_layer_removal}
    \end{minipage}\hfill
    \begin{minipage}[t]{.58\textwidth}
    \caption{Left: Outgoing weights from the last convolutional layer are re-ordered by the flatten layer. Only one channel is depicted. Right: We mimic max pooling with stride 2, kernel size 2 on the outgoing weights of the last convolutional layer. White blocks are discarded.}
    \label{fig:3d}
    \end{minipage}
\end{figure}

Unfortunately SVD cannot be easily extended towards convolutional layers.
Instead we look at available pruning techniques.
Since we do not wish to set individual connecting weights to zero, but actually remove all incoming and outgoing weights of for example a convolutional channel, we need to apply a structured pruning technique, see Figure \ref{fig:cnn_layer_removal}.
As discussed in Section \ref{sec:related}, such methods often require expensive search algorithms, a large amount of re-training, or both; neither of which is feasible as each operation for ECToNAS which will get performed hundreds of times.

\citet{Liu2020} propose a pruning technique for convolutions based on the parameters of the following batch normalisation (BN) layer given by
\begin{equation}
    y = \gamma \frac{x - \mu}{\sqrt{\sigma^2 + \epsilon}} + \beta\label{eq:batch_norm},
\end{equation}
with input $x$, $\mu$ and $\delta$ representing the mean and standard deviation of the current mini-batch, scaling factor $\gamma$ and offset factor $\beta$.
There is one $\gamma$ and $\beta$ in a BN layer per channel in the previous convolutional layer.
\citet{Liu2020} argue that a small $\gamma$ means that the corresponding channel has little influence on the layer's total output, and can thus be set to zero.
The affected channel's $\beta$ value is added to the bias term of the following layer.
We adopt this idea and thus always pair convolutional with BN layers.
When removing channels from a convolutional layer, we identify and cut away the ones corresponding to the smallest $\gamma$ values, while adding the $\beta$ term to the following layer's bias term.
Similarly, removing a whole convolutional layer means cutting out the whole layer channel by channel.
This type of operation greatly affects the overall input-output behaviour, and the weights of the newly generated child network are thus at best a warm start for further training, however we still find that this technique produces acceptable results and works as a trade off between computational cost and not having to completely re-train from scratch.
In \citet{Liu2020}, the L1-penalty of $\gamma$ is added to the loss function in order to promote pruning capabilities.
Since this is not our main intent, we skip this term.

This leaves us with one more operation that we wish to include, namely pooling.
Pooling layers work fundamentally different than all other layers we have discussed so far, as they have no weights of their own and cannot be initialised to maintain overall input-output behaviour of the network.
In terms of implementation, modifying a pooling layer does not affect the weights of successive convolutional layers (which are independent of respective input shapes), instead we need to look towards the first subsequent dense layer, which may be many steps down the line.
We find that aggregating the input weights to the first dense layer as if we had applied the relevant pooling operation (that means using the same stride and aggregation function, see Figure \ref{fig:3d}) produces acceptable results and can again be seen as a warm start towards re-training.
A more detailed explanation is included in Appendix \ref{app:pooling}.
Conversely, when removing a pooling layer we simply copy the incoming connections to the dense layer such that applying the pooling operation again would yield the original weights.

Adding convolutional layers without pooling drastically increases the overall network parameter count.
Since we already always pair convolutional with BN layers, we instead define convolutional cells as fixed elements that may be added or removed only as a whole.
Such convolutional cells are thus made up of the following combination (and order) of layers: convolution - pooling - batch normalisation - activation.
Finally this gives us the pool of allowed network operations:
\begin{itemize}
    \item Identity
    \item Add/remove dense layer
    \item Add/remove units to dense layer
    \item Add/remove convolutional cell
    \item Add/remove channels
\end{itemize}

These operations define the theoretical search space from which ECToNAS can identify winning topologies.
In practice, the search space is further restricted by the provided starting topology and available computational budget.

\subsection{Architecture Evaluation}\label{sec:evaluation}
Choosing a good balance between greedily optimising for validation accuracy or also rewarding smaller networks is not a trivial problem, as the optimal solution is not clearly defined and heavily dependent on available computational resources and task specifics.
While validation accuracy has a clearly defined range of [0, 1] with 1 being the ultimate goal, there exist no hard limits nor a ubiquitously desirable optimum on network parameter count.
Similarly when regarding loss instead of accuracy we not only have to change the goal towards decreasing the metric, but also keep in mind that achievable loss does not have a universal upper bound either.

In its fitness function, \emph{the Surgeon} \citep{Schiessler2021} weighs validation accuracy $v$ against a function of accuracy gain $\Delta v$ and parameter fraction $\Delta p$ (both calculated with respect to a candidate's parent), amounting to $v + \exp(\Delta v)/\exp(\Delta p)$.
This is based on \citet{Cai2018a}'s rationale that `a 1\% increase in validation accuracy should yield a higher gain if it occurs from 90\% to 91\% instead of 60\% to 61\%'.
We find that the decision of how much focus to place on each aspect cannot be made globally and therefore make it dependent on a user choice parameter $\alpha \in [0, 1]$.
ECToNAS thus has two modes: greedy ($\alpha = 1$) and balanced ($\alpha < 1$).
The resulting fitness score $s$ is given by
\begin{equation}
    s = \begin{cases}
        v & \alpha = 1\\
        \alpha \Delta v - (1-\alpha) \Delta p & \alpha < 1
    \end{cases}
\end{equation}
where again $\Delta v$ and $\Delta p$ are computed wrt. the candidate's parent.
Note that competition phase 1 in Algorithm \ref{alg:ectonas} is always evaluated greedily, and the balanced version is used in phase 2 if desired.

\section{Experiments}
We choose six different starting topologies for ECToNAS, three of which already contain one or more convolutional cells. 
The rest are simple feed forward neural networks of various sizes.
Both the FFNNs and the topologies already containing one or more convolutional cells are further categorised by sizes (`small', `medium' or `large'), amounting to a total of 6 possible combinations.
We perform 10 runs on each starting topology, with a computational budget of 1,000 total epochs per run for purely greedy mode ($\alpha = 1$) as well as in two non-greedy variants ($\alpha = 0.5$ and $\alpha = 0$).
For our baseline we use standard training for 200 epochs performed on the same starting topologies.
Experimental results are reported in detail for a fixed random seed and repeated twice more using different random seeds to confirm overall trends and reduce potential deterministic effects.
Detailed descriptions of the starting topologies and further hyperparameters are included in Appendix \ref{app:hyperparams}.

As an additional ablation study we deactivate the decision score and instead randomly select the winner in each competition bracket, which we call random mode.

Our experiments are performed across several standard image classification data sets:
Cifar 10 \& 100 \citep{Krizhevsky2009}, Eurosat \citep{Helber2018, Helber2019}, Fashion MNIST \citep{Xiao2017}, MNIST \citep{LeCun1998}, SVHN \citet{Netzer2011}.
Details on these data sets can be found in Appendix \ref{app:datasets}.
10\% of each training set are set aside as validation data, and a further 10\% as test data in case no separate test split is available.
Training and validation data are used by ECToNAS, the set aside test sets are only used for final evaluation purposes.
ECToNAS keeps track of validation accuracy during training and uses these scores in the fitness function, as is discussed in section \ref{sec:evaluation}.

In Table \ref{tab:results_overall} we compare the mean test accuracies over all runs per data set achieved by ECToNAS on greedy setting (E, $\alpha = 1$) with the baseline (B) and random mode (R).
On the test set, ECToNAS outperforms the baseline in all cases, whereas the random mode scores similarly or slightly worse than the baseline with a small to moderate increase in average parameter count of the final architectures.
Note that the parameter count of the baseline corresponds to that of the starting topologies, as these are not changed during standard neural network training.

\begin{table}[htb]
    \centering
    \begin{tabular}{l|rlrlrl|rrr}
         & \multicolumn{6}{c|}{Test set accuracy [\%]}  & \multicolumn{3}{c}{Parameter count}\\
        \diagbox{Data\\set}{Mode} & \multicolumn{2}{c}{E} & \multicolumn{2}{c}{B} & \multicolumn{2}{c|}{R} & \multicolumn{1}{c}{E} & \multicolumn{1}{c}{B} & \multicolumn{1}{c}{R}\\
        \hline
        CIFAR-10 & 54.2 &$\pm$ 3.4& 46.3 &$\pm$ 4.7 & 47.1 &$\pm$ 3.5 & 100 K & 77 K & 74 K\\
        CIFAR-100 & 20.6 &$\pm$ 3.1 & 16.0 &$\pm$ 1.9 & 14.8 &$\pm$ 3.8 & 97 K & 78 K & 80 K\\
        EuroSAT & 78.1 &$\pm$ 10.5 & 70.8 &$\pm$ 20.0 & 67.6 &$\pm$ 15.2 & 301 K & 303 K & 283 K\\
        FashionMNIST & 87.2 &$\pm$ 1.5& 86.8 &$\pm$ 1.4 & 85.3 &$\pm$ 2.7 & 43 K & 28 K & 29 K\\
        MNIST & 98.3 &$\pm$ 1.6 & 97.6 &$\pm$ 2.2 & 96.0 & $\pm$ 3.4& 43 K & 28 K & 29 K\\
        SVHN & 77.2 &$\pm$ 5.1 & 66.4 &$\pm$ 13.0 & 66.6 &$\pm$ 10.2 & 68 K & 77 K & 77 K\\
        \hline
        overall & 69.3 &$\pm$ 26.0 & 64.0 &$\pm$  28.7 & 62.9 &$\pm$  27.6 & 109 K & 99 K & 95 K
    \end{tabular}
    \caption{Performance on the test set and parameter count of final architecture of greedy ECToNAS (E, $\alpha = 1$) compared to unmodified baseline (B) and random mode (R) for various data sets. Data are aggregated over all different starting topologies.}
    \label{tab:results_overall}
\end{table}

\begin{figure}[htb!]
    \centering
    \begin{minipage}{.48\textwidth}
    \includegraphics[width=\textwidth]{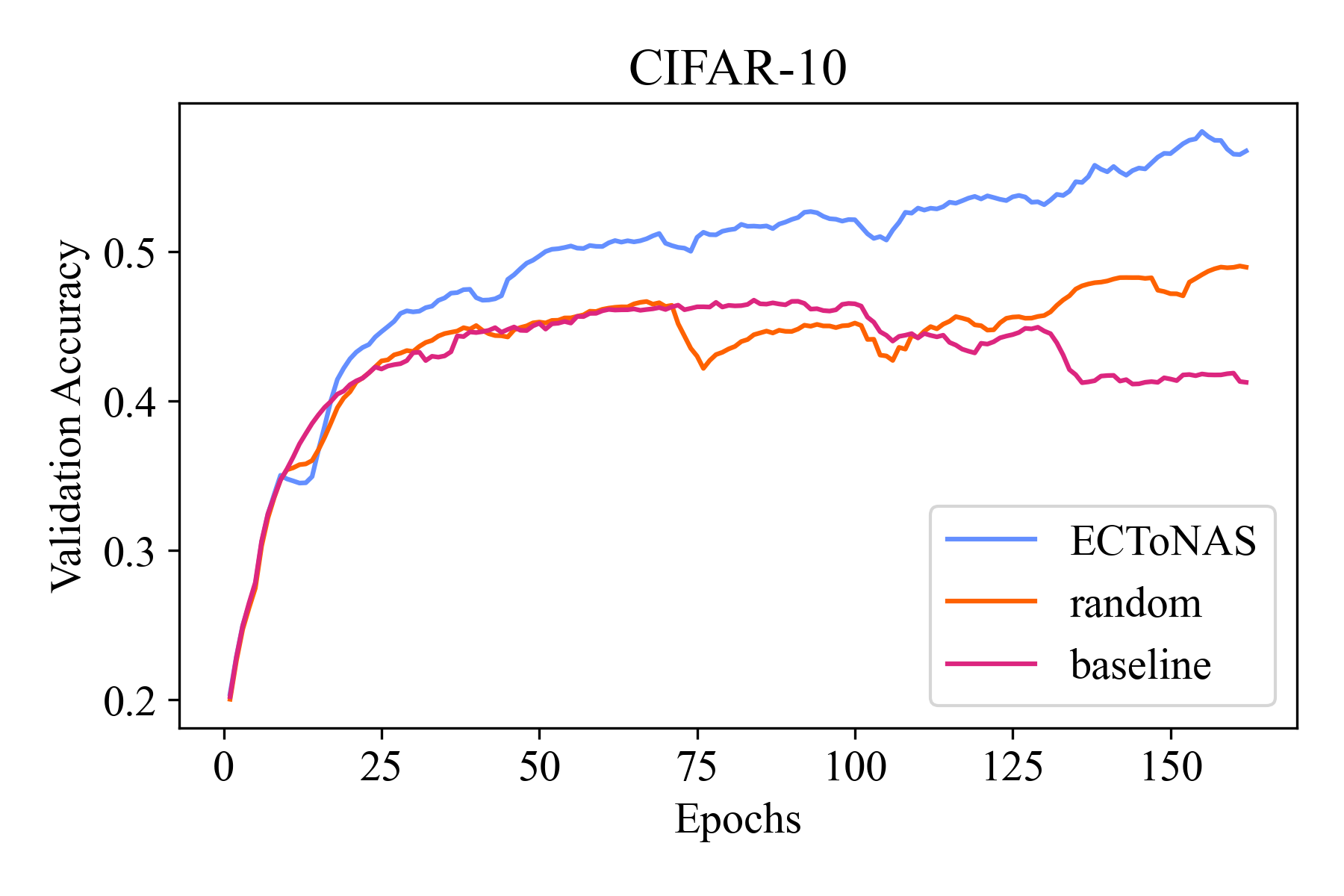}
    \end{minipage}\hfill
    \begin{minipage}{.48\textwidth}
    \includegraphics[width=\textwidth]{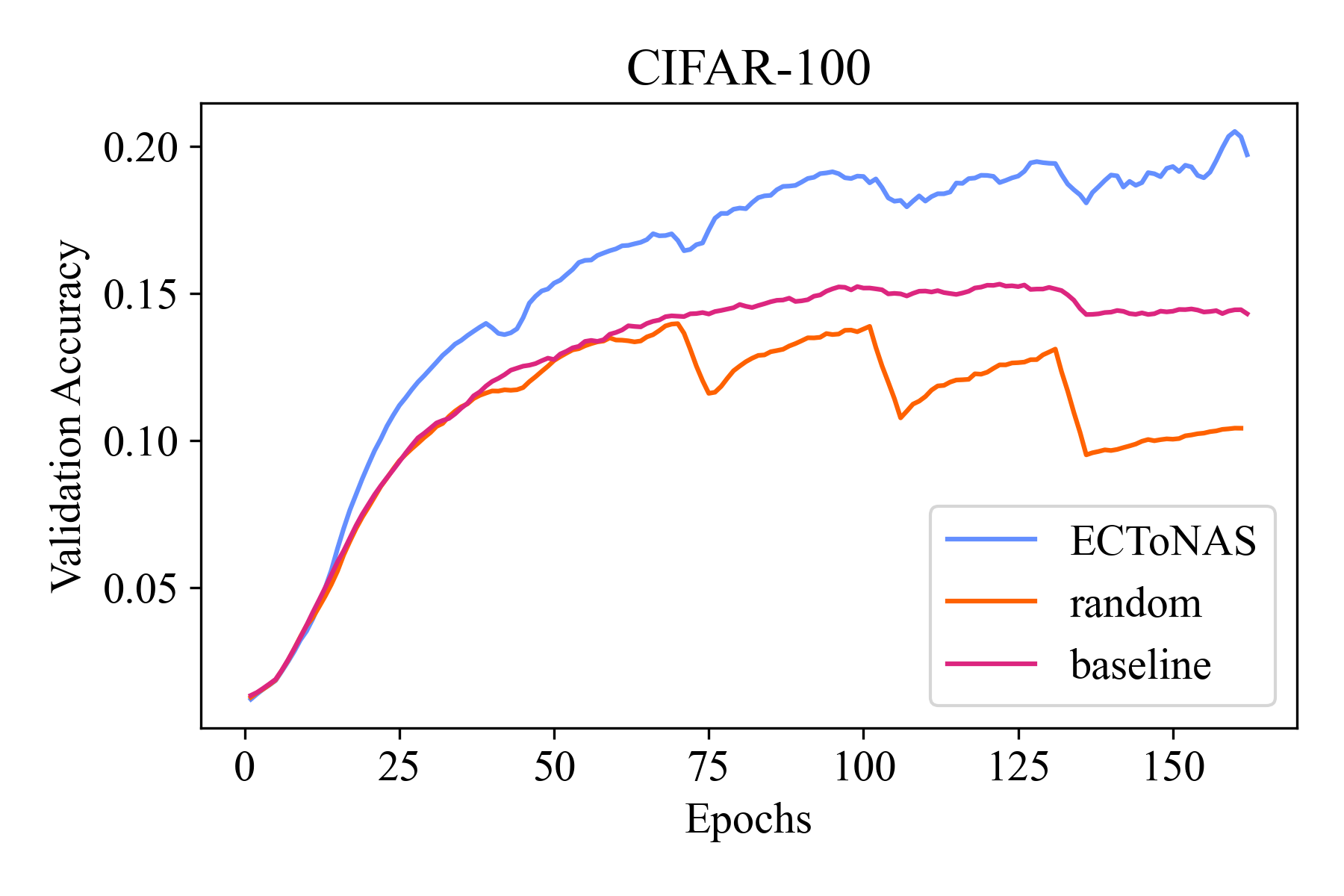}
    \end{minipage}\hfill
    \begin{minipage}{.48\textwidth}
    \includegraphics[width=\textwidth]{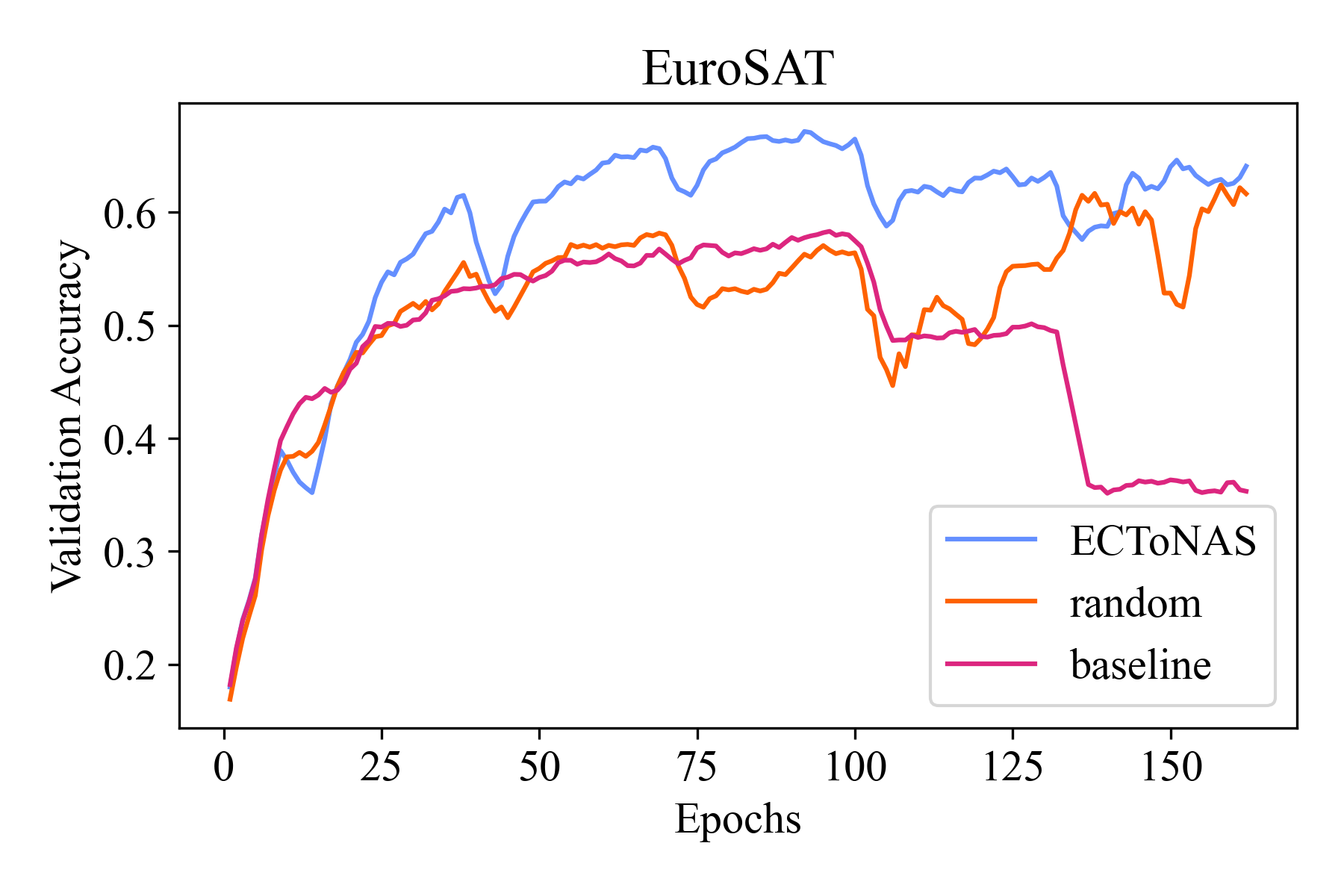}
    \end{minipage}\hfill
    \begin{minipage}{.48\textwidth}
    \includegraphics[width=\textwidth]{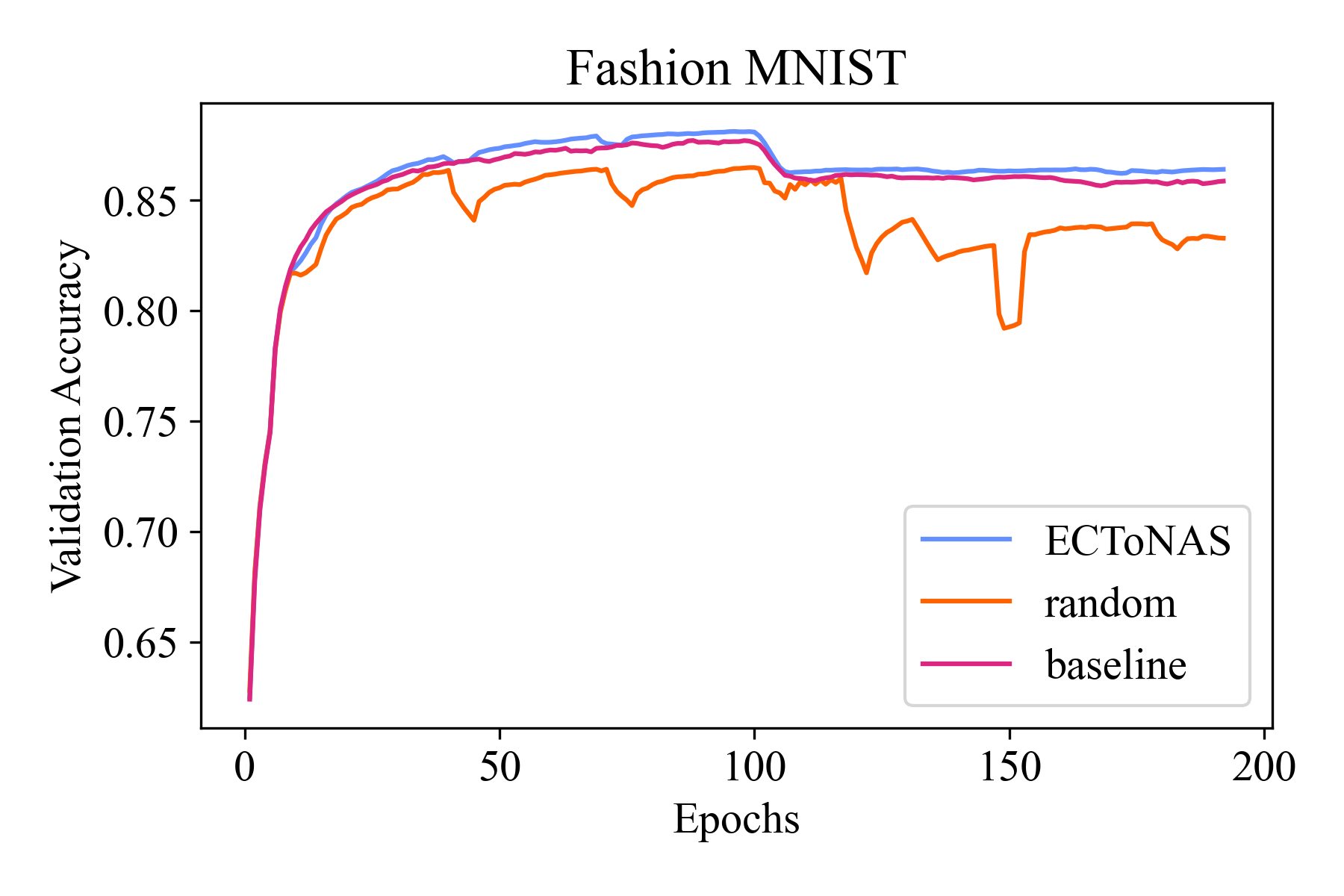}
    \end{minipage}\hfill
    \begin{minipage}{.48\textwidth}
    \includegraphics[width=\textwidth]{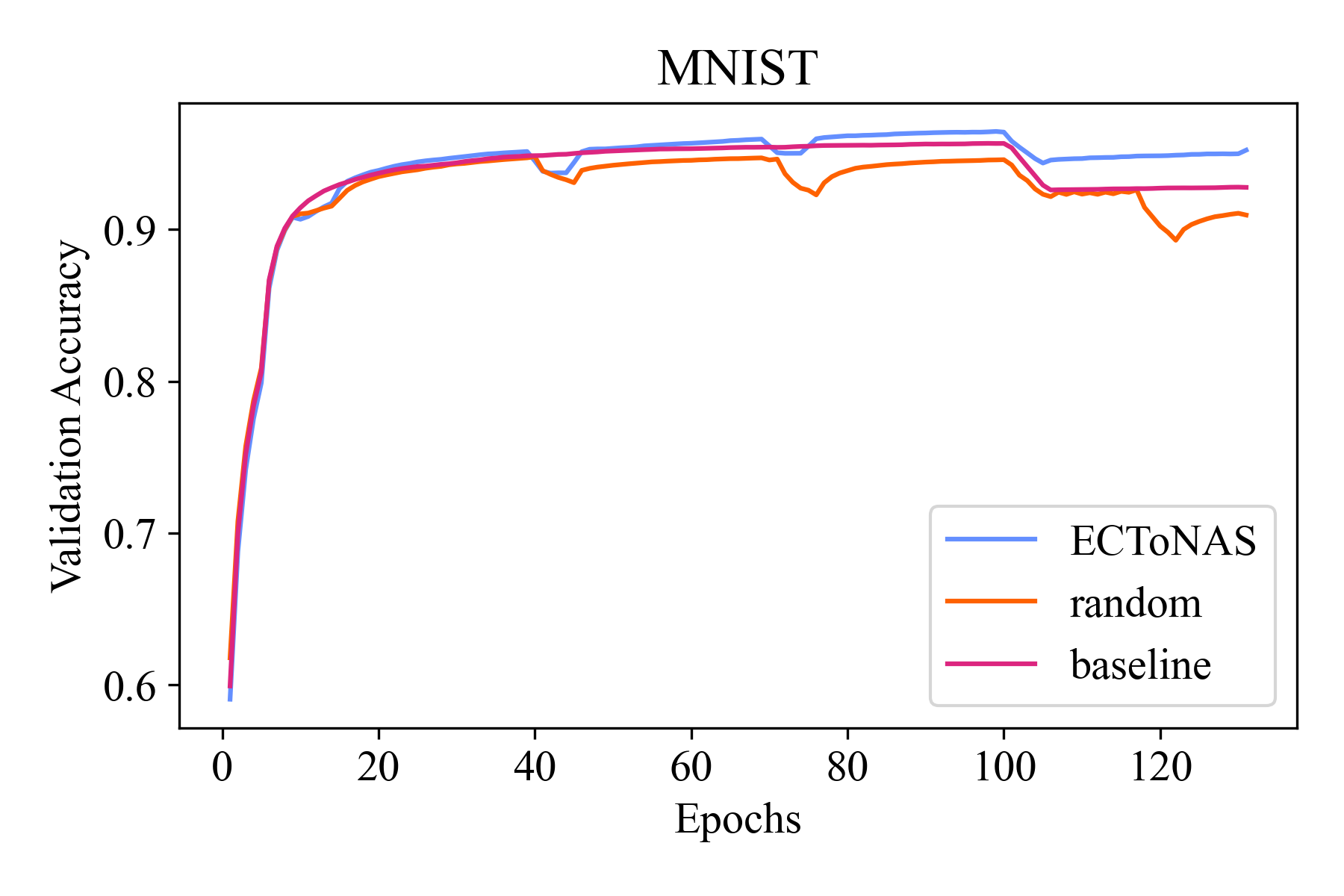}
    \end{minipage}\hfill
    \begin{minipage}{.48\textwidth}
    \includegraphics[width=\textwidth]{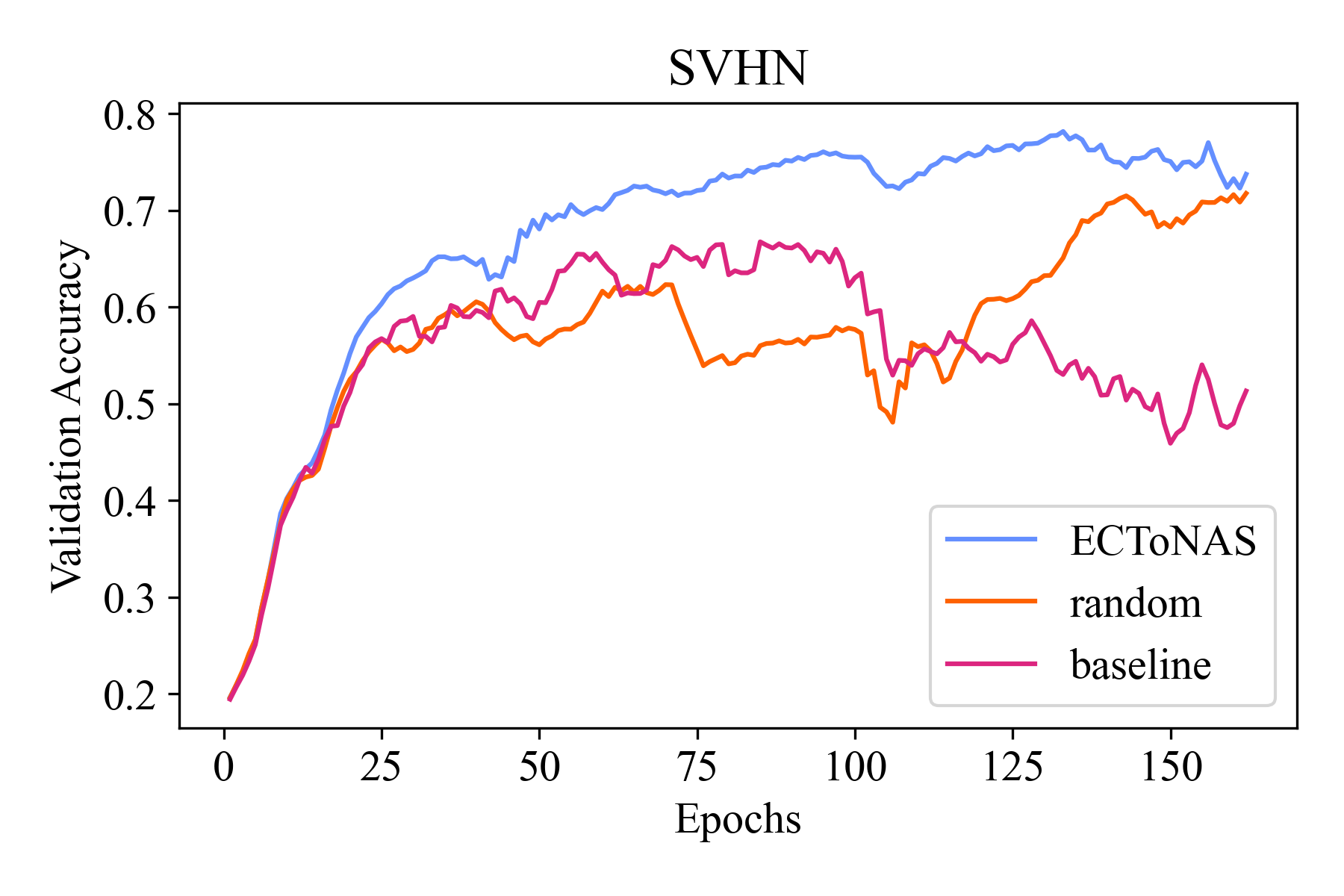}
    \end{minipage}
    \caption{Comparison of accuracy on the validation set achieved by ECToNAS (greedy mode, $\alpha = 1$) with random mode and unmodified baseline topology across different data sets. Graphs are aggregated over all starting topology variations.}
    \label{fig:results_overall}
\end{figure}

By construction, each network candidate created by ECToNAS remembers the training that its parent has received, with as small memory losses as possible in case of drastic structural changes.
Thus, the parent's count of received epochs of training are also included in the child's training count, such that the final optimised architecture produced by ECToNAS may have received between 100 to 160 epochs of training in total.
When compared to re-training each candidate from scratch, as most NAS algorithms do, this means that ECToNAS can save up to around 80\% of training time while still producing a fully trained final neural network.
ECToNAS is limited by overall computational budget to be spread out over all generated candidates, thus we cannot directly control the number of epochs that the winner will receive.
This number is influenced by a number of factors such as the size and structure of the initial topology and various hyperparameter settings which control for example the number of candidates generated and the maximum size of each parent generation.
We see this difference when averaging over runs with different starting topologies for the same data set, where sudden jumps in otherwise relatively flat validation accuracy curves may occur due to some runs ending earlier than others.
Figure \ref{fig:results_overall} depicts average validation accuracies achieved by ECToNAS (greedy mode, $\alpha = 1$) compared to the native baseline and random mode.
To make comparison fairer, we cut off the data for baseline and random mode for each starting topology after the same amount of epochs that ECToNAS on average spent on its winning topology.
Apparent sudden jumps in accuracy levels (which do not recover within a few epochs) stem from averaging over runs with such varying cut-off points, such as for example the graph for the baseline trained on the EuroSAT data set in Figure \ref{fig:results_overall}, left hand side, middle row.

In Table \ref{tab:ecto_greedy_levels} we compare results generated using ECToNAS with different greediness settings ($\alpha \in \{1, 0.5, 0\}$).
In the greedy version ($\alpha = 1$), ECToNAS as expected beats the non-greedy versions wrt. final test accuracies by around 5-10 percentage points.
It is noteworthy however that the non-greedy settings yield drastically compressed networks (achieving compression rates between 80\% and 90\%), and still in almost all cases manage to score at least as well as the baseline.

\begin{table}[htb]
    \centering
    \begin{tabular}{l|rlrlrl|rrr}
         & \multicolumn{6}{c|}{Test set accuracy [\%]} & \multicolumn{3}{c}{Parameter count}\\
        \diagbox{Data\\ set}{$\alpha$} & \multicolumn{2}{c}{1} & \multicolumn{2}{c}{0.5} & \multicolumn{2}{c|}{0} &
         \multicolumn{1}{c}{1} & \multicolumn{1}{c}{0.5} & \multicolumn{1}{c}{0}\\
         \hline
        CIFAR-10 & 54.2 & $\pm$ 3.4 & 49.8 & $\pm$ 5.1 & 50.0 & $\pm$ 5.1 &100 K & 8 K & 8 K\\
        CIFAR-100 & 20.6 & $\pm$ 3.1 & 18.2 & $\pm$ 3.3 & 15.0 & $\pm$ 2.6&97 K & 7 K & 8 K\\
        EuroSAT & 78.1 & $\pm$ 10.5 & 71.3  & $\pm$ 7.1 & 70.7 & $\pm$ 8.7&301 K & 36 K & 17 K\\
        Fashion MNIST & 87.2 & $\pm$ 1.5 & 83.7 & $\pm$ 2.3 & 83.9 & $\pm$ 2.6&43 K & 6 K & 6 K\\
        MNIST & 98.3 & $\pm$ 1.6 & 94.7 & $\pm$ 3.2 & 94.8 & $\pm$ 3.2&43 K & 6 K & 6 K\\
        SVHN & 77.2 & $\pm$ 5.1 & 69.7 & $\pm$ 9.2 & 70.5 & $\pm$ 11.7&68 K & 14 K & 9 K\\
        \hline
        overall & 69.3 & $\pm$ 26.0 & 64.6 & $\pm$ 25.5 & 64.1 & $\pm$ 26.8& 109 K & 13 K & 9 K
    \end{tabular}
    \caption{Performance on the test set and parameter count of final architecture compared over various ECToNAS greediness levels ($\alpha$). Data are aggregated over all different starting topologies.}
    \label{tab:ecto_greedy_levels}
\end{table}

\begin{figure}[htb]
    \centering
    \begin{minipage}{.48\textwidth}
    \includegraphics[width=\textwidth]{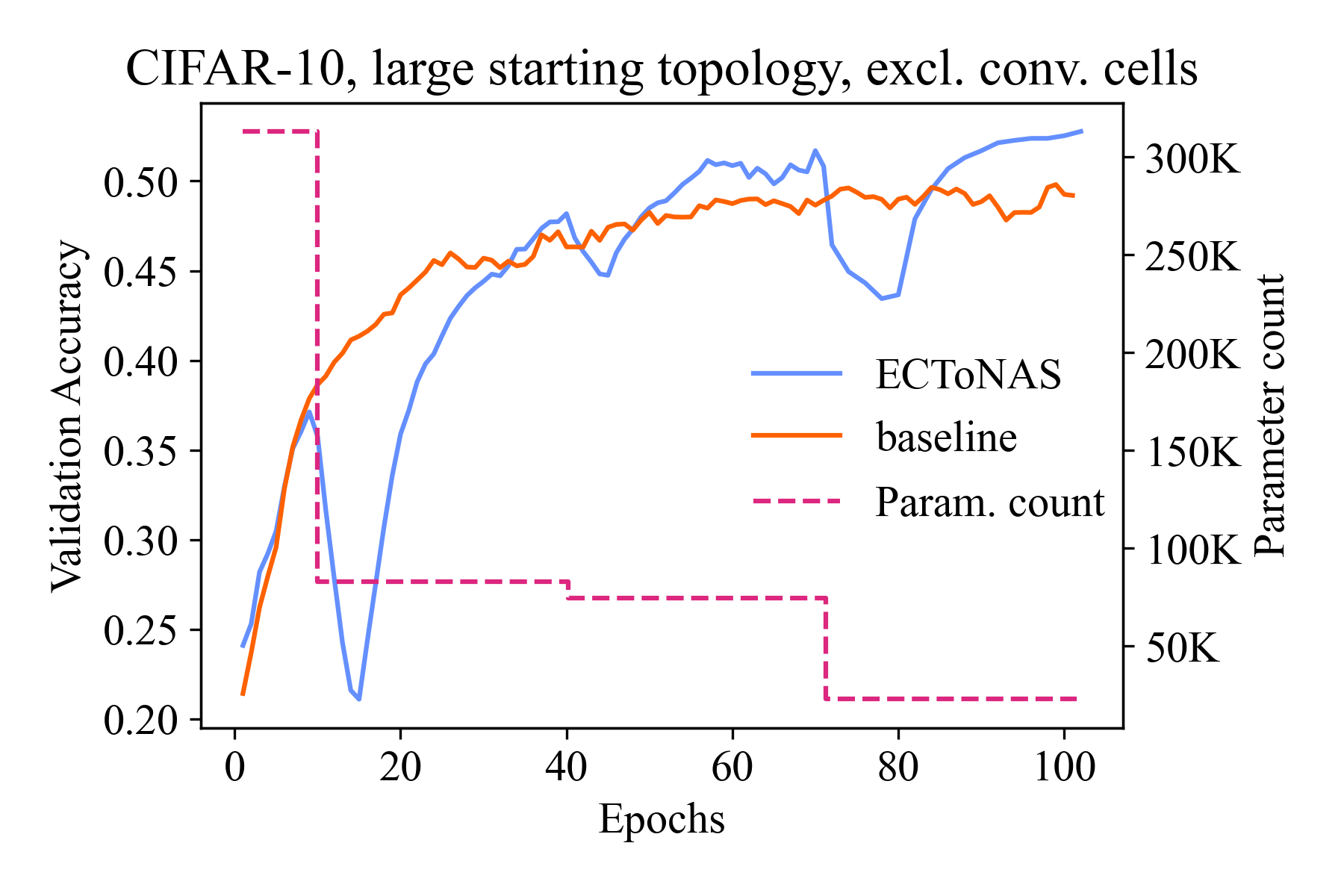}
    \end{minipage}\hfill
    \begin{minipage}{.48\textwidth}
    \includegraphics[width=\textwidth]{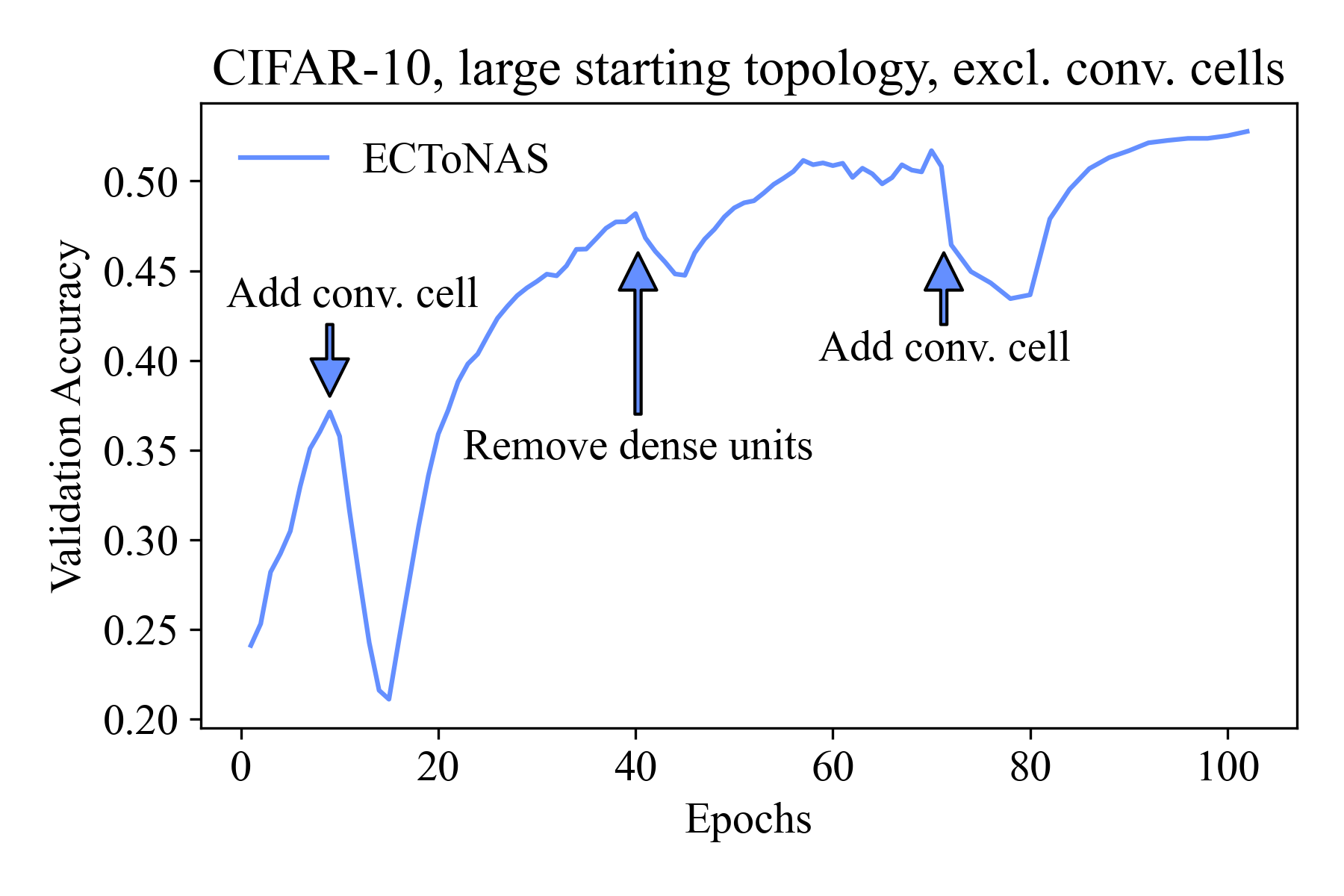}
    \end{minipage}
    \caption{Example of a single run of non-greedy ECToNAS ($\alpha = 0.5)$ on the CIFAR-10 data set starting off from a large topology without convolutional cells present.
    Left: Comparison with naive baseline and development of network parameter count during training.
    ECToNAS manages reach a higher validation accuracy level while simultaneously compressing the network to around 7\% of its original size.
    Right: Modifications performed by ECToNAS are annotated.}
    \label{fig:cifar10_detail}
\end{figure}

An example summarizing a single run of non-greedy ECToNAS ($\alpha = 0.5$) is presented in Figure \ref{fig:cifar10_detail}.
We can see ECToNAS's performance wrt. to both validation accuracy and network parameter count (left hand figure) as well as annotations describing which modifications were selected by the algorithm (right hand figure).

All our experiments are performed and reported for a fixed random seed (see Appendix \ref{app:hyperparams}).
In order to confirm behaviour and limit the influence of unwanted deterministic effects we repeated all experiments twice more on different random seeds.
These results fall in line with what has been presented.

The wall times per experimental run using the chosen data sets and starting topologies vary between 10 and 45 minutes, with most runs completing in around 20-30 minutes.

\subsection{Convolutions on Tabular Data Set}
Starting from a small initial topology and expanding layers or even introducing new topological elements as required is considered to be the most common use case for ECToNAS.
However, it also has the ability to remove existing convolutional layers all together, leaving a fully connected feed forward neural network instead.
This is handy when owing to unclear data type and objective function or simple lack of experience, a user might train a network with with one or several convolutional layers included that are not required or even harmful in order to achieve good results.

Since to the best of our knowledge there exist no well-known benchmark data sets for this use case, we artificially generate data that are structured like images (meaning tabular data will be formatted to look like $n \times m$ `pixels'), but where there is no structural information contained.
We base this off the well known adult data set \citep[also known as census data set,][see also Appendix \ref{app:datasets}]{Kohavi1996}.

We one-hot encode any ordinal values, and drop any \texttt{nan}-columns, as well as the columns `sex-Male' and the three native countries with the lowest count, in order to arrive at 100 columns.
Numerical values are min-max-scaled into the range [0, 1].
Finally the column order is scrambled, and the data rearranged to mimic $10 \times 10$ `pixel' image data.

Using this artificially constructed data set, we run ECToNAS again on all six starting topologies that are described in Appendix \ref{app:hyperparams} and Table \ref{tab:starting_topos}.
We perform 25 runs with various greediness levels ($\alpha \in \{0, 0.5, 1\}$), as well as random and naive baseline mode.
Thus 75 runs in total start off from either a CNN or a FFNN respectively for each mode.
Again the greedy version of ECToNAS ($\alpha = 1$) manages to outperform all other modes with regard to test set accuracy.
The final parameter count compared to the baseline goes down in case of the non-greedy versions, and up for greedy ECToNAS and random mode.
Table \ref{tab:adult_results} details these results and is included in the appendix.

For this experiment we are however more interested in the nature of the winning topology.
Specifically we want to see how often topology crossing happens \emph{away} from CNNs towards FFNNs.
We expect to see few if any topology crossings towards FFNNs in non-greedy versions of ECToNAS, as the fitness function in these cases rewards smaller networks.

\begin{table}[htb]
    \centering
    \begin{tabular}{l|r r r|r r r}
        Starting topology  & \multicolumn{3}{c|}{CNN} & \multicolumn{3}{c}{FFNN} \\
        Final topology & CNN & FFNN & Crossing & CNN & FFNN & Crossing\\
        \hline
        Greedy ($\alpha = 1$) & 70 & 5 & 6.7\% & 0 & 75 & 0.0\% \\
        Non-greedy ($\alpha = 0.5$) & 75 & 0 & 0.0\% & 67 & 8 & 89.3\%\\
        Non-greedy ($\alpha = 0$) & 75 & 0 & 0.0\% & 74 & 1 & 98.6\%\\
    \end{tabular}
    \caption{Counts of resulting topology types per starting topology type and greediness level, as well as percentage of runs that ended in topology crossing.}
    \label{tab:adult_crossing}
\end{table}

Indeed, in Table \ref{tab:adult_crossing} we see this expected behaviour confirmed.
Non-greedy versions of ECToNAS never cross towards a FFNN, as this would lead to increase in network size, and in most cases do cross towards CNN for the same reason.
Greedy ECToNAS instead never crosses towards CNN, and in a few cases indeed removes all initially present convolutional cells (in addition to any that might have been added at run time) to end up with a standard FFNN.
This can be seen as proof of concept that topology crossing indeed happens both ways when using ECToNAS.

\section{Conclusion}
We presented ECToNAS, a lightweight, computationally cheap evolutionary algorithm for cross-topology neural architecture search.
One of its advantages over other neural architecture search algorithms is ECToNAS's ability to re-use network weights when generating new candidate networks, by manipulation of existing, pre-trained weights.
To accomplish this we introduced several techniques for weight modification that aim to minimise the change in input-output behaviour caused by adding, removing, or changing structural elements of a neural networks' topology.

We showed that ECToNAS is able to supersede naive training, or reach comparable validation accuracy while being able to considerably compress the size of a given starting topology.
The user is able to influence the desired outcome by tuning a greediness parameter, an can thus control how much focus should be placed on achievable validation scores vs. network parameter count.
Ablation studies that deactivated its selection function demonstrated that ECToNAS is able to successfully identify those network candidates with the greatest potential.

ECToNAS tends to favour adding convolutional cells (a fixed sequence of the following layers: convolution, pooling, batch normalisation, activation) over removing them.
The addition of the pooling layer shrinks network parameter count, which is rewarded in the fitness function, whereas the batch normalisation layer helps to stabilise the learning process.
We analysed and showed that topology crossing towards a FFNN, that is removal of all initially present convolutional cells, does happen as well under certain circumstances.

There are still some weaknesses inherent to this algorithm which were deemed to be out of scope for this manuscript but can be extended upon in future works.
First and foremost, in its current state ECToNAS is restricted to sequential networks only.
A possible starting point for how to get rid of this limitation may be found in \citet{Cai2018a}.
There the authors show how to apply their Net2Net operations (which serve as basis for some methods we presented) to DenseNet, for example when using add operations.
This limits achievable accuracy levels, and forces a somewhat outdated structure of any produced network architecture that is not comparable to the in general much more complex state-of-the-art models.

Secondly, ECToNAS is currently running on a very strict cyclical scheme.
Mutation and selection phases are scheduled after a fixed amount of training, regardless of any internal states of the network candidates such as convergence rate etc.
Introducing a dynamic learning rate may also serve to improve validation accuracy results.
Early stopping could be used to further reduce computational requirements, but since the non-greedy version of ECToNAS at times selects candidates that perform worse wrt. their validation accuracy, it is somewhat unclear what criteria for early stopping could be applied that do not interrupt the algorithm too early.

Reaching state-of-the-art predictive accuracy levels is not everyone's goal however.
Researchers without a strong background in machine learning often struggle to choose an appropriate architecture for their models, while potentially also having a limited availability of training data and/or computational resources.
ECToNAS is ideally suited for these use cases since it can expand or shrink provided topologies as required and provides a fully trained, ready to use final neural network.

ECToNAS should thus be seen as a proof of concept that cross-topology neural architecture search is possible, and is intended to serve as a starting point for further research.

\section*{Acknowledgements}
This research did not receive any specific grant from funding agencies in the public, commercial, or not-for-profit sectors.

\appendix

\section{Further Details on Network Operations}
Here we will provide further details for network operations that either have already been published previously in \citet{Chen2016} and \citet{Schiessler2021}, or would go into too much detail in the main manuscript body.

\subsection{Fully Connected Layers}\label{app:fully_connected}
\citet{Chen2016} introduce Net2DeeperNet and Net2WiderNet operations that can be used to add fully connected layers or widen existing layers.

When adding a new layer, its weight matrix is initialised as identity matrix and fully trainable.
Under at least piecewise linear activation functions such as \texttt{relu}, this preserves the overall network output:
Let $\phi$ be an activation function, then 
\begin{equation}
    \phi(v) = \phi(\mathcal{I}_d \phi(v))
\end{equation}
for all vectors $v$ and identity operation $\mathcal{I}_d$, iff $\phi$ is at least piecewise linear.

Widening existing layers is done by copying and scaling some of the existing weights.
Let fully connected layers $L_i$ and $L_{i+1}$ have weight matrices $W^{(i)} \in \R^{m\times n}$ and $W^{(i+1)} \in \R^{n \times p}$.
The Net2WiderNet operation allows replacing $L_i$ by a different layer $L_i^*$ that has $q > n$ outgoing connections.
For this, \citet{Chen2016} define a random mapping function $g:\{1, 2, \dots, q\} \rightarrow \{1, 2, \dots, n\}$ such that
\begin{equation}
    g(j) = \begin{cases}
    j & j \leq n \\
    \text{random sample from }\{1, 2, \dots, n\} & j > n
    \end{cases}
\end{equation}
The weight matrices $W^{(i)}$ and $W^{(i+1)}$ are replaced by $U^{(i)}$ and $U^{(i+1)}$ with
\begin{equation}
    U^{(i)}_{k,j} = W^{(i)}_{k,g(j)}, ~ U^{(i+1)}_{j,h} = \frac{1}{|\{x|g(x)=g(j)|\}} W^{(i+1)}_{g(j),h}
\end{equation}
The scaling factor is introduced to ensure that all units in the modified network have the same value as in the original network, even though weights may have been copied several times.
Since we only applied linear operations, the same argument as above still holds true that Net2WiderNet preserves overall network output under at least piecewise linear activation.

As trimming information will in general change the overall input-output behaviour, \emph{the Surgeon} \citep{Schiessler2021} makes use of singular value decomposition (SVD) to remove neurons from a fully connected layer, or even whole layers:
Given an arbitrary matrix $A \in \R^{m\times n}$ we can use SVD to find a representation
\begin{equation}
    A = U\Sigma V^T
\end{equation}
with orthogonal $U \in R^{m\times m}, V \in \R^{n\times n}$ and rectangular diagonal $\Sigma \in \R^{m\times n}$ with $k\leq \min(m, n)$ non-negative real entries $\sigma_i$ along its diagonal.
By convention, the $\sigma_i$ (called singular values) are given in descending order of magnitude.
Reducing the rank of $A$ to $r < k$ is done by setting $\sigma_i = 0,~ \forall i> r$ and dropping associated rows and columns of $U$ and $V$.

Thus, $U\Sigma = \tilde{A} \in \R^{m\times r}$, with $\tilde{A}$ being the closest representation of $A$ with rank $r$ under Frobenius norm.
$\tilde{A}$ can now serve as a new weight matrix for a dense layer with $r$ units instead of $n$.
In order to reconnect this layer to the following one, the subsequent weight matrix needs to be multiplied from the left with $V^T$.

In \citet{Schiessler2021} we note that it is not possible to accommodate for effects of even linear activation functions using this method; thus the activation function is simply ignored and changes in input-output behaviour have to be mitigated with further training.
Thus \emph{the Surgeon} removes whole fully connected layers in a similar fashion by multiplying their weights onto the weight matrix of the subsequent layer, which preserves incoming and outgoing dimensions and ignores any present activation functions.

\subsection{Pooling Layers}\label{app:pooling}
\begin{figure}[htb]
    \centering
    \begin{minipage}{.495\textwidth}
    \includegraphics[width=.9\textwidth]{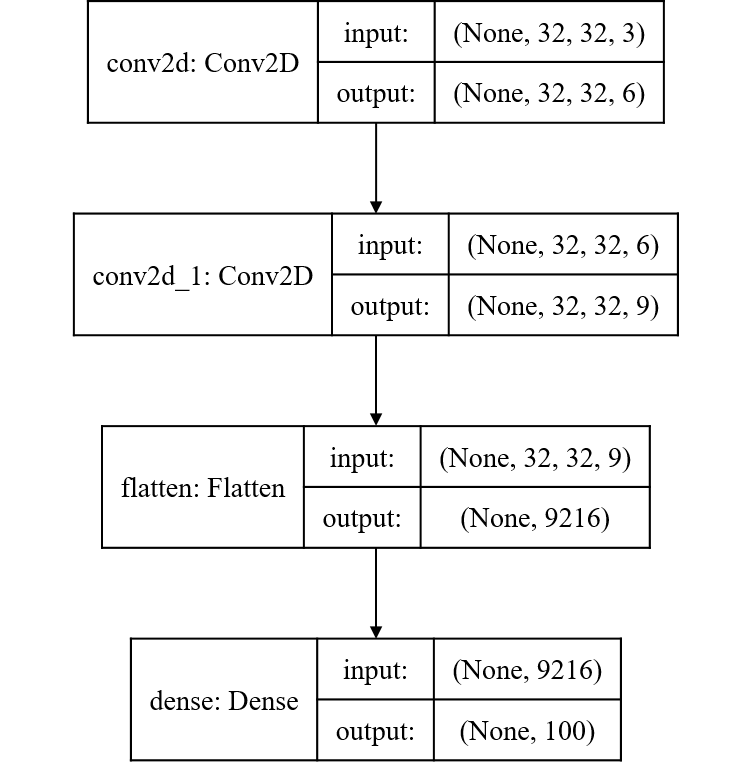}
    \end{minipage}\hfill
    \begin{minipage}{.495\textwidth}
    \includegraphics[width=.9\textwidth]{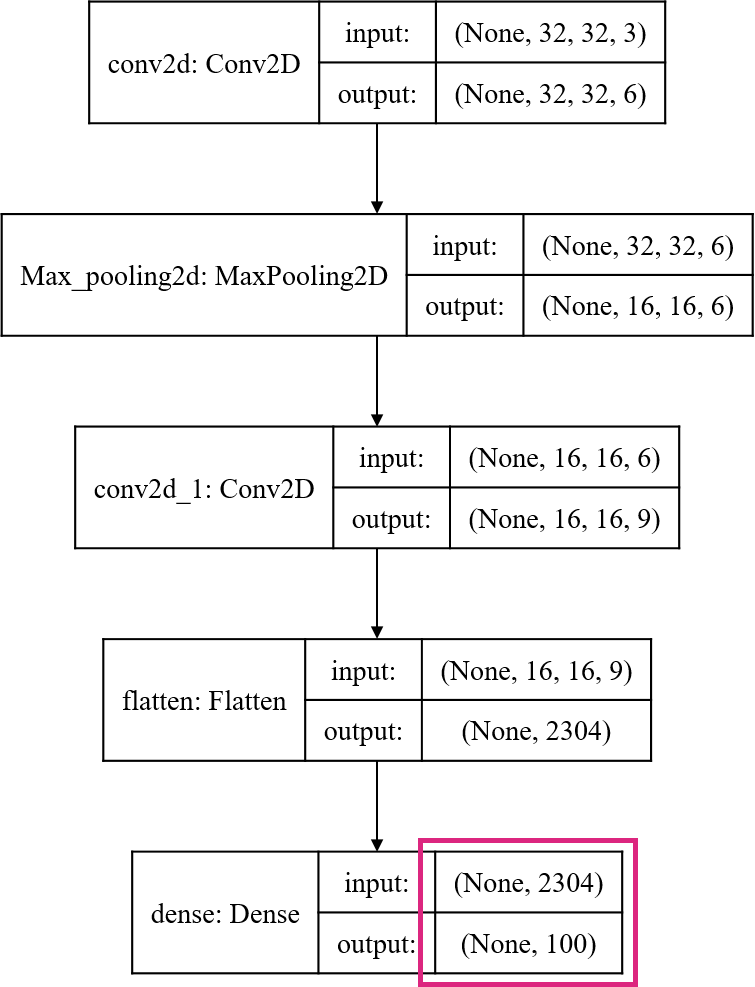}
    \end{minipage}
    \caption{Introducing a pooling layer affects input and output shapes of all subsequent layers up to the first dense layer, where the weight matrix now has to accommodate for a different incoming dimension.}
    \label{fig:tf_pool_structure}
\end{figure}

Pooling layers are structurally different from dense or convolutional layers in that they have no trainable weights of their own.
As such, when we introduce a new pooling layer into an existing, pre-trained network, we cannot initialise the layer in such a way that it behaves as (close as possible to) an identity operation in terms of overall input-output behaviour.
The second way in which the introduction of a pooling layer differs from other layer types is that this has structural consequences not only locally (meaning on the layer itself and its direct successor), but also further downstream up until the first dense layer, see Figure \ref{fig:tf_pool_structure}.

In particular, any convolutional or other layers in between the newly introduced pooling layer and the first dense layer also see the difference in input shapes, however their weights are not affected.
Since we cannot retain (spatial) information that was lost through pooling, we opt to not concern ourselves with the affected convolutional layers and simply allow successive re-training to make up for lost accuracy.
The same cannot be done with the first dense layer, which needs to have its weight matrix reduced to (w.l.o.g., depending on pooling layer settings and rounding effects) around $1/4^{th}$ of its original shape.
A very simple approach would be to just randomly cut away superfluous input connections to the dense layer.
Structured trials (discussed in \ref{app:pooling_trials}) have shown however that we can actually retain more information by mimicking the effect of pooling on the incoming connections to the flatten layer.
Note that the flatten layer simply reshapes incoming information into vector format, such that the dense layer can parse it without any further altercations, see Figure \ref{fig:3d} (left).

In the example provided in Figure \ref{fig:tf_pool_structure}, the flatten layer without pooling receives each training sample along a $(32 \times 32 \times 9)$ pixel weight matrix, or synonymously along 9 channels of $(32 \times 32)$ pixel weight matrices.
We then introduce a max pooling layer with kernel size 2, stride 2 and zero-padding as required.
The very top left $2\times 2$ pixels of the first channel of each input sample have now been passed through max pooling, retaining only the single pixel which carries the most information.
Which of the four pixels is selected will be different for each sample, and there is no way of knowing or controlling that.
We reason however that picking the outgoing weight of this $2\times 2$ area with the largest absolute value has the greatest potential of retaining as much information as possible.
The same of course applies to all other pixel groups in each sample.

Thus, when introducing a pooling layer, we apply the same aggregation function that is used for pooling (maximum or average) to each $2 \times 2$ weight block of the flatten layer input (using padding where required), and form a new weight matrix of the appropriate shape $(16 \times 16 \times 9)$, which in turn is flattened to 2304 input connections.

Conversely, removal of pooling layers is done by repeating each incoming connection to the flatten layer the appropriate amount of times, and inserting them at the correct position.
This works for both max and average pooling.

\subsubsection{Pooling Layer Trials}\label{app:pooling_trials}
To evaluate the soundness of the proposed pooling method we ran structured trials on a minimal example network consisting of two convolutional followed by two dense layers, which we trained for 25 epochs using the MNIST data set.
This baseline network achieved an average validation accuracy of 98.6\% over 100 different random seeds (resulting in different train-test-validation splits of the data set).
We then inserted a pooling layer between the two convolutional layers and compared the loss in validation accuracy when adapting the weights using our proposed pooling method vs. randomly cutting away existing connections.
The modified networks then each received one epoch of re-training, and validation accuracies where compared again.
Results are summarised in Table \ref{tab:pooling}.

\begin{table}[htb]
    \centering
    \begin{tabular}{l|r r | r r}
         Pooling method & \multicolumn{2}{c|}{Structured} & \multicolumn{2}{c}{Random} \\
         Retraining & none & 1 epoch & none & 1 epoch \\
         \hline
         Average pooling & 89.1\% & 96.0\% & 53.8\% & 95.6\% \\
         Max pooling & 13.4\% & 29.7\% & 49.3\% & 95.0\% 
    \end{tabular}
    \caption{Achieved validation accuracies of structured pooling trials using the MNIST data set on a minimal example network. Results are averaged over 100 runs using different train-test-validation splits. The baseline achieved an average validation accuracy of 98.6\%.}
    \label{tab:pooling}
\end{table}

From these trials we can see that the proposed method works really well in case of average pooling, but fails in case of max pooling, where it surprisingly achieves significantly worse results than just randomly cutting away connections.
After one epoch of re-training, all but the structured max-pooling versions have regained a significant amount of the lost validation accuracy, whereas the max pooling layers are still quite far behind.
We surmise that the proposed method shows promise especially when introducing average pooling layers where little retraining can be provided, and may be insufficient in combination with max pooling layers.
Since we only tested on one data set and one example topology, there may well be cases where our method works also in the case of max pooling.

ECToNAS is able to independently select which type of pooling layer to apply when introducing new convolutional cells.
Even with this type of pooling adaptation, we see cases where max pooling cells get chosen.
Our proposed method of introducing pooling layers should thus be seen as a starting point for further investigations.

\section{Additional Results}\label{app:additional_results}
In this section we include further results that were omitted from the main body of the manuscript.
Table \ref{tab:adult_results} contains detailed results from experiments performed on the modified adult set.
Note that the adult set has a very high class imbalance where the instances in both training as well as test set are split approximately 76\%/24\% across the two classes.
This needs to be taken into account when comparing predictive results.

\begin{table}[htb]
    \centering
    \begin{tabular}{l l|c|c}
         Mode & Greediness & Test set accuracy [\%] & Parameter count\\
         \hline
        Baseline & - & 84.4 $\pm$ 0.4 & 5.5K\\
        \hline
         & greedy ($\alpha = 1$) & 84.8 $\pm$ 0.7 & 7.6K\\
        ECToNAS & non-greedy ($\alpha = 0.5$) & 83.6 $\pm$ 1.2 & 1.5K\\
         & non-greedy ($\alpha = 0$) & 83.5 $\pm$ 1.3 & 1.4K\\
        \hline
        Random & - & 83.7 $\pm$ 1.2 & 6.5K
    \end{tabular}
    \caption{Performance on the test set and parameter count of final architecture of the modified adult data set using different greediness settings for ECToNAS compared to random mode and naive baseline. Data are aggregated over all different starting topologies.}
    \label{tab:adult_results}
\end{table}

\section{Implementation Details}\label{app:implementation_details}
In this section we provide details on data sets and our implementation of ECToNAS.
The underlying code can be found at \url{https://github.com/ElisabethJS/ECToNAS}.

\subsection{Data Sets}\label{app:datasets}
All image data sets were downloaded from the tensorflow data sets catalogue available at \url{https://www.tensorflow.org/datasets/catalog/overview} (last accessed on 2022-09-06).
The tabular Adult data set was downloaded from the UCI Machine Learning repository \citep{Dua2017} at \url{http://archive.ics.uci.edu/ml/datasets/Adult} (last accessed on 2022-09-06).

Adult \citep{Kohavi1996}:
Census income data set.
Tabular data set containing 48,842 samples of categorical and numerical data across 14 attributes.
Some missing values occur.
Can be used to predict whether a person's income will be below or above USD 50K\$/year.
This data set has a heavy class imbalance.

Cifar 10 \& Cifar 100 \citep{Krizhevsky2009}:
Images from the Canadian Institute of Advanced Research.
$(32 \times 32)$ px colour images that are evenly divided into 10 or 100 classes depending on the version.
50,000 training and 10,000 test images are available.

Eurosat \citep{Helber2018, Helber2019}:
Sentinel-2 satellite images.
$(64 \times 64)$ px images. We use the RGB Version that has 3 colour channels and consists of 10 classes.
27,000 samples are available.

Fashion MNIST \citep{Xiao2017}:
Data set of the fashion company Zalando's article images.
$(28 \times 28)$ px grey-scale images evenly belonging to 10 categories.
60,000 training and 10,000 test images are available.

MNIST \citep{LeCun1998}:
Modified National Institute of Standards and Technology data set of handwritten digits.
$(28 \times 28)$ px grey-scale images evenly belonging to 10 classes that represent the digits 0-9.
60,000 training and 10,000 test images are available.

SVHN \citep{Netzer2011}:
Images of house numbers from Google Street View.
We use a cropped version of $(32 \times 32)$ px colour images that evenly fall into 10 classes representing the numbers 0-9.
73,257 training and 26,032 test images are available, as well as 531,131 extra training samples which are not used.

\subsection{Hyperparameter Settings}\label{app:hyperparams}
We make use of a total of 6 different starting topologies which are summarised in Table \ref{tab:starting_topos}.
We differentiate between them by their size (with options `small', `medium' or `large'), and by whether or not any convolutional cells are included.
Convolutional cells consist of a CNN layer (with number of channels specified in Table \ref{tab:starting_topos}), followed by an average pooling, batch normalisation and activation layer.
We use \tf's \emph{padding='same'} option for both convolutional and pooling layers, and standard stride and kernel sizes, as well as \texttt{relu} activations.
The dense layers also use \texttt{relu} activations, with respective unit count specified in Table \ref{tab:starting_topos}.

\begin{table}[htb]
    \centering
    \begin{tabular}{l | c|c c}
        Size & CNN cells included & conv. channels & dense units \\
        \hline
        small & true & 3, 6, 9 & 10\\
         & false & - & 10\\
        \hline
        medium & true & 3, 6 & 10, 10\\
         & false & - & 10, 10\\
        \hline
        large & true & 3 & 100, 50, 10\\
         & false & - & 100, 50, 10
    \end{tabular}
    \caption{Number of convolutional channels and dense units present in the various starting topologies used for ECToNAS experiments.}
    \label{tab:starting_topos}
\end{table}

Note that the size descriptors relate to parameter count of the resulting neural network, which is why a topology that includes three convolutional cells (and thus a total of three pooling layers) is considered to be smaller than a topology that only contains one such cell.
Each starting topology thus further consists of an input layer and a further dense output layer with \texttt{softmax} activation, with input shape and number of output channels tuned to the respective data set.

ECToNAS can be run with a number of different hyperparameters.
Notable amongst these are the computational budget, greediness weight $\alpha$ and random seed, as well as indirect hyperparameters such as training and test set and optimizer, loss function and scoring metric.

For our experiments we chose the following settings:
\begin{itemize}
    \item Computational budget: 1,000 epochs
    \item Greediness weight: 0, 0.5 and 1
    \item Random seed: 42 (extra experiments run on seeds 13 and 28)
    \item Optimizer: stochastic gradient descent, with learning rate 0.1
    \item Loss: sparse categorical cross entropy
    \item Scoring metric: accuracy
\end{itemize}
Further parameters can be set via code and are described in the documentation.

\subsection{Python Environment Requirements}
We used the following packages and versions to run ECToNAS.
Important dependencies are marked with an asterisk.
\texttt{
\begin{itemize}
    \item python 3.8 *
    \item tensorflow 2.3.0 *
    \item keras 2.4.0 *
    \item numpy 1.21.5
    \item pandas 1.4.2
    \item re 2.2.1
\end{itemize}
}

\subsection{Hardware specifications}
Our experiments were performed on a virtual machine running on a 24-core 2.1 GHz Intel Xeon Scalable Platinum 8160 processor, which is equipped with a Tesla V100 GPU card with 16 GB memory.

\section{NAS Best Practice Checklist}
When performing our experiments and writing up this manuscript we adhered to \emph{the NAS Best Practices Checklist} (Version 1.0.1, available at \url{https://www.automl.org/nas_checklist.pdf}) as published by \citet{Lindauer2020}.

\bibliographystyle{nat_adapt}
\bibliography{publications_extended}
\end{document}